\documentclass[11pt]{article}

\usepackage[final]{acl}

\usepackage{times}
\usepackage{latexsym}
\usepackage[T1]{fontenc}
\usepackage[utf8]{inputenc}
\usepackage{microtype}
\usepackage{inconsolata}
\usepackage{graphicx}
\usepackage{booktabs}
\usepackage{multirow}
\usepackage{amsmath}
\usepackage{amssymb}
\usepackage[table,dvipsnames]{xcolor}  % "table" for \rowcolor; "dvipsnames" enables ForestGreen used in the Delta row
\usepackage{enumitem}
\usepackage{xspace}
\usepackage[most]{tcolorbox}
\usepackage{fvextra}  % extends fancyvrb with breaklines / breakanywhere
\fvset{breaklines=true,breakanywhere=true,fontsize=\footnotesize,xleftmargin=0pt,xrightmargin=0pt}
\definecolor{promptHdr}{HTML}{4A4A4A}
\definecolor{promptBdy}{HTML}{F4F4F4}
\definecolor{promptVar}{HTML}{2C7DA0}
\newtcolorbox{promptbox}[1]{%
  enhanced,
  colback=promptBdy, colframe=promptHdr,
  coltitle=white, fonttitle=\bfseries\small,
  title={#1}, halign title=flush left,
  arc=1.2mm, boxrule=0pt,
  left=2.5mm, right=2.5mm, top=2mm, bottom=2mm,
  before skip=2pt, after skip=2pt,
  fontupper=\normalsize\raggedright,
}

\newcommand{\pv}[1]{\textcolor{promptVar}{\texttt{\{#1\}}}}
\newcommand{\cspg}{\textsc{SAP}\xspace}
\newcommand{\Pinit}{\ensuremath{P_{i}}\xspace}
\newcommand{\Poutput}{\ensuremath{P_{o}}\xspace}
\newcommand{\Pcreate}{\ensuremath{P_{c}}\xspace}
\newcommand{\Pumsg}{\ensuremath{P_{u}}\xspace}
\newcommand{\Pfallback}{\ensuremath{P_{f}}\xspace}

\title{SAP: State-Guided Data Synthesis with Argument Provenance \\for Multi-Turn Tool Use}
\author{%
  \textbf{Zichen Tian}\textsuperscript{1},
  \textbf{Jinpeng Chen}\textsuperscript{1}\thanks{Project lead},
  \textbf{Cheng Gong}\textsuperscript{2},
  \textbf{Suiyun Zhang}\textsuperscript{2},
  \textbf{Rui Liu}\textsuperscript{2}\thanks{Corresponding author.} \\
  \textsuperscript{1}Independent Researcher \quad
  \textsuperscript{2}Huawei Research \\
  \texttt{zichentian1024@outlook.com, jinpeng.chen@my.cityu.edu.hk, liu.rui2@huawei.com}
}

\begin{document}
\maketitle

% =====================================================================
\begin{abstract}
High-quality multi-turn tool-use data is essential for training agentic models, yet existing data synthesis methods often underrepresent the argument-level dependencies that are critical to long-horizon tool use. As a result, even when a model selects the correct tool, task execution may still fail because the model fills tool arguments with fabricated, stale, or weakly grounded values. To address this problem, we propose \textbf{State-Guided Data Synthesis with Argument Provenance (\cspg{})}. \cspg{} combines state guidance, tool-argument provenance constraints, and turn-level validation to efficiently construct tool-use trajectories with long-range dependencies and high accuracy. Using data generated by \cspg{}, we build \cspg-4B, which is highly competitive even when compared with much larger models across multiple benchmarks. Source code, synthesized data, and trained weights are available at \url{https://github.com/Zichen1024/SAP}.
\end{abstract}
% =====================================================================

\section{Introduction}
\label{sec:intro}

Tool use is the primary interface through which large language models (LLMs) interact with the real world~\citep{patil2025bfcl, yao2024taubench, barres2025tau2bench}. As LLMs are deployed in long-horizon agentic systems, multi-turn tool use becomes a key axis of agent capability; yet the complexity of state evolution, tool feedback, and cross-turn dependencies makes high-quality multi-turn trajectories scarce, a major bottleneck for training agentic models~\citep{rabinovich2025robustness, zhang2025mtasurvey, prabhakar2025apigenmt}.

\begin{figure}[t]
\centering
\includegraphics[width=\linewidth]{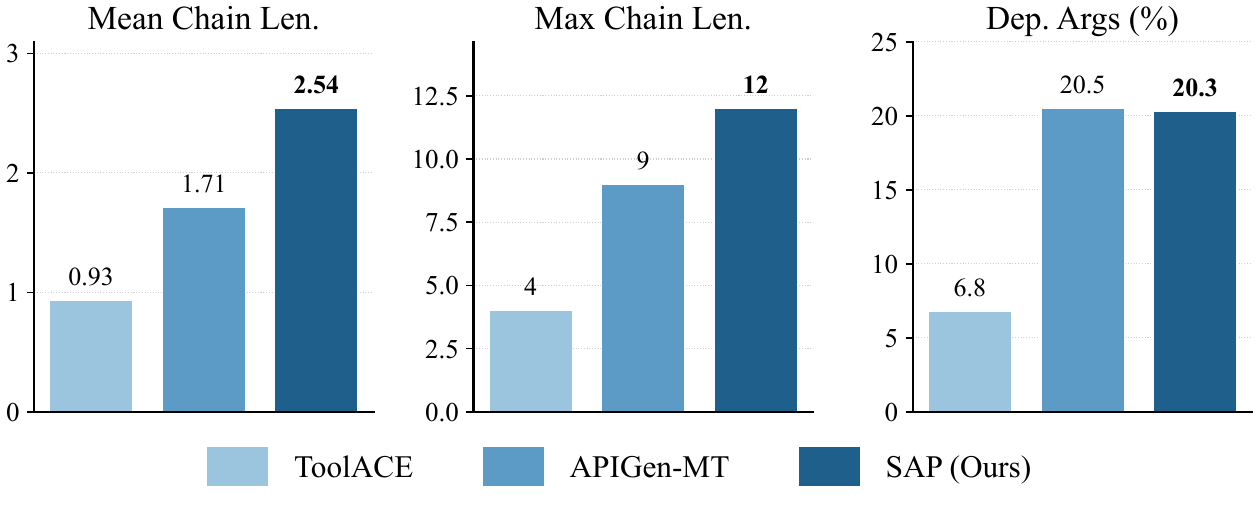}
\caption{Argument-dependency statistics on representative open-source multi-turn tool-use training sets and ours. \emph{Mean Chain Len.} and \emph{Max Chain Len.} denote the mean and maximum lengths of argument-value propagation chains across turns; \emph{Dep. Args (\%)} denotes the proportion of arguments involved in cross-turn dependencies.}
\label{fig:ppg-comparison}
\end{figure}

In practice, many multi-turn failures arise not from choosing the wrong tool, but from incorrectly specified tool arguments~\citep{rabinovich2025robustness}. Yet existing data-synthesis work mostly focuses on tool-selection dependencies: MAGNET~\citep{yin2025magnet} and FunReason-MT~\citep{xu2025funreasonmt}, for example, use pre-built tool dependency graphs to ensure tool-sequence validity, while cross-turn dependencies among tool arguments are largely overlooked. Such dependencies are ubiquitous; arguments typically come from the initial environment state, prior tool-call returns, or earlier user messages, and when underrepresented in training data, the resulting models fabricate upstream references, reuse stale user inputs, or hallucinate substitutes when an upstream call fails. To quantify this gap, we measure the length of cross-turn argument-value propagation chains and the proportion of arguments involved in such dependencies, and find that existing open-source training datasets~\citep{liu2024toolace, prabhakar2025apigenmt} are consistently shallow on both,\footnote{Sample sizes for the statistics in Figure~\ref{fig:ppg-comparison}: ToolACE 11k, APIGen-MT 5k, \cspg{} (Ours) 9k.} leaving models prone to errors in complex multi-turn tasks~\citep{rabinovich2025robustness}.

Beyond provenance, existing methods also face a hard trade-off between trajectory accuracy and generation cost. Two paradigms currently dominate. \textbf{Intent-First} methods~\citep{qin2024toolllm, prabhakar2025apigenmt, xu2025toucan, zeng2026toolacemt, chen2026covetraininginteractivetooluse, xu2026gem} draft user intents first and expand dialogues and tool calls post hoc; their validation typically combines rule-based blueprint checks (schema, types, executability) with LLM committee review. Owing to the limited reliability of the underlying LLMs, this validation cannot fully guarantee accuracy, and a single failed check often discards the entire trajectory even when it contains locally correct sub-sequences. \textbf{Trace-First} methods~\citep{yin2025magnet, xu2025funreasonmt, hao2026hardgen} sample executable trajectories over a tool-dependency graph and derive user-side queries from those traces; this gives strong executability guarantees but couples the graph to the source tool ecosystem, making migration to new domains costly.

To address these issues, we propose \textbf{State-Guided Data Synthesis with Argument Provenance (\cspg{})}. Given a toolset and its documentation, \cspg{} dynamically constructs a Finite State Machine (FSM) whose edges already carry per-argument provenance tags; it then samples state-transition trajectories from the FSM and fills tool arguments turn by turn. Arguments that depend on prior context are validated against the executed history, and each tool call is executed immediately after generation. If a call fails, only the offending call is retried rather than discarding the trajectory. Once the call sequence is fixed and verified, the framework generates the corresponding natural-language user and assistant messages. By combining state-transition guidance, parameter-provenance awareness, and immediate tool-call validation, \cspg{} jointly supports executability, causal consistency of arguments, and cross-domain scalability, achieving both high quality and efficiency.

Using data synthesized by \cspg{}, we train \cspg-4B, which achieves competitive results on both BFCL v4 multi-turn~\citep{patil2025bfcl} and $\tau^2$-bench Retail/Airline~\citep{barres2025tau2bench}. To support future research, the synthesis code, synthesized data, and trained weights are publicly available at \url{https://github.com/Zichen1024/SAP}. Our main contributions are as follows:
\begin{itemize}
  \item We introduce argument-provenance diagnostics for multi-turn tool-use data, showing that existing open-source data contain shallow dependency chains and few arguments that depend on previous turns, limiting their support for learning long-range argument tracking.
  \item We propose \cspg{}, a state-guided synthesis framework that dynamically builds an FSM from tool documentation, treats argument provenance as an explicit constraint, and validates tool calls turn by turn, enabling the efficient construction of multi-turn tool-use data with strong causal dependencies and high correctness.
  \item We show that \cspg-4B achieves competitive results on both BFCL v4 multi-turn and $\tau^2$-bench; ablation studies further confirm the importance of each component.
\end{itemize}

% ===================================================================

\section{Related Work}
\label{sec:related}

\paragraph{Multi-turn tool-use data synthesis.}
High-quality multi-turn trajectory synthesis is a principal bottleneck for tool-augmented LLMs~\citep{patil2025bfcl, yao2024taubench, barres2025tau2bench, prabhakar2025apigenmt, xu2025funreasonmt, guan2025kgrag, hao2026hardgen, 2025inftool, 2026covert, chai2026uikobe, 2026failurereflect}. The \textbf{intent-first} paradigm starts from user intent and expands tool invocations post hoc, instantiated as DFS planning over a large API pool~\citep{qin2024toolllm}, structured blueprints validated by rule-based checks and LLM committees~\citep{prabhakar2025apigenmt}, non-autoregressive skeletons with mask-and-fill refinement~\citep{zeng2026toolacemt}, graph-sampled planned generation for dialogue coherence~\citep{wang2024toolflow}, user-side intent modeling at scale~\citep{2026useroriented}, scaling tool-use synthesis from real-world MCP environments~\citep{xu2025toucan}, or distillation of implicit tool-use intents from raw text~\citep{xu2026gem}. The \textbf{trace-first} paradigm samples executable traces over a tool dependency graph (hand-curated, signature-derived, or execution-evolved) and derives user queries conditioned on those traces~\citep{yin2025magnet, xu2025funreasonmt, hao2026hardgen, 2026astra}. A recent line of \emph{provenance-aware planning}, exemplified by ToolWeave~\citep{khandelwal2026toolweave}, constructs tools with built-in dependencies and tracks parameter provenance at plan time to reduce argument hallucination.

In contrast, \cspg{} extends the type system to five categories (\Pinit, \Poutput, \Pcreate, \Pumsg, plus the runtime-only \Pfallback) and requires every declared source to be resolved against the executor state prior to binding, with no pre-built or evolved dependency graph, enforcing argument-level causal consistency at synthesis time rather than post hoc. Along a similar line, ToolMind~\citep{2025toolmind} applies per-turn filtering after generation to catch errors that propagate across turns, whereas \cspg{} lifts the check from a post-hoc filter into a synthesis-time constraint. InfTool~\citep{2025inftool} co-evolves synthesized data and the trained model via outer-loop GRPO, whereas \cspg{} keeps the executor loop within a single synthesis pass and produces fixed SFT data.

\paragraph{Multi-turn tool-use benchmarks.}
Several benchmarks target multi-turn tool use. BFCL v4~\citep{patil2025bfcl} is among the most widely adopted; its multi-turn track is partitioned into base, long-context, miss-param, and miss-func subsets that probe distinct capabilities. The $\tau$-bench family~\citep{yao2024taubench, barres2025tau2bench} constructs state-aware Retail/Airline dialogues, requiring the agent to maintain a coherent world state under an LLM-simulated user over long horizons. ToolDial~\citep{2025tooldial} and DialogTool~\citep{2025dialogtool} provide complementary multi-turn / stateful tool-use datasets. \citet{rabinovich2025robustness} expose the fragility of function calling under distribution shift, and \citet{zhang2025mtasurvey} survey multi-turn LLM interactions more broadly.

% =====================================================================

\begin{figure*}[t]
  \centering
  \includegraphics[width=1.0\linewidth]{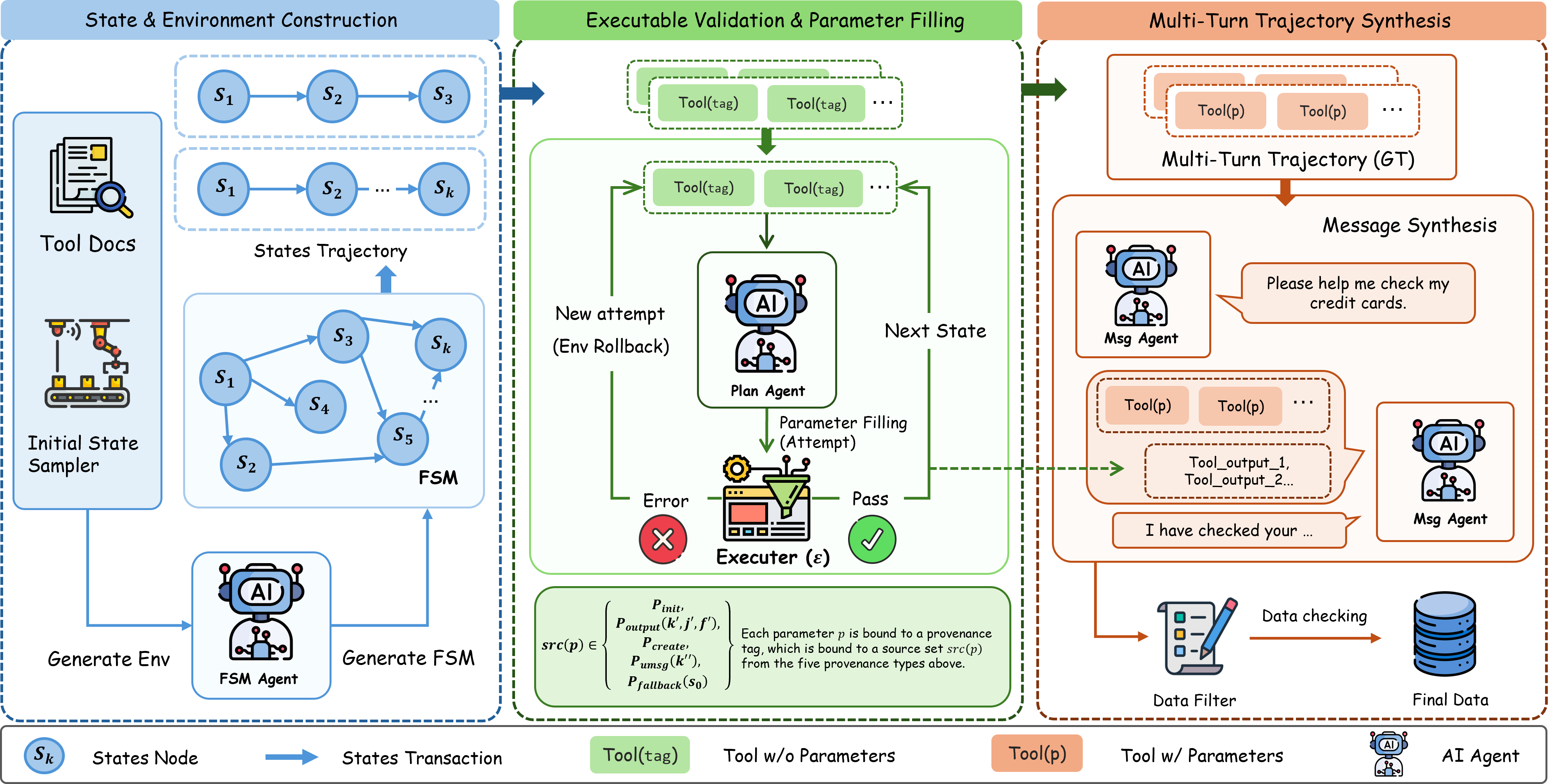}
  \caption{Overview of the \cspg{} pipeline. From left to right, $\mathcal{A}_{\text{FSM}}$ constructs an FSM from tool documentation and sampled initial states, and the pipeline samples a state trajectory from it. $\mathcal{A}_{\text{plan}}$ fills the tagged calls $\mathrm{Tool}(\mathrm{tag})$ with concrete parameters, yielding $\mathrm{Tool}(p)$, and executes them with $\varepsilon$, using environment rollback and local retry on failure. Finally, $\mathcal{A}_{\text{msg}}$ synthesizes the messages for the validated trajectory, which is filtered into the final data. Each argument is assigned a provenance source from $\{\Pinit,\Poutput,\Pcreate,\Pumsg,\Pfallback\}$.}
  \label{fig:pipeline}
\end{figure*}

% =====================================================================
\section{Preliminaries}
\label{sec:prelim}

\subsection{Task Definition}
We formalize multi-turn tool-use data synthesis as a \textbf{Partially Observable Markov Decision Process (POMDP)}, defined by the tuple $\mathcal{M} = (S, A, O, T)$, where $S$ is the latent environment state space (the world state of the tool-execution environment, e.g., database contents or session context); $A = A_{\text{tool}} \cup A_{\text{resp}}$ is the action space, with $A_{\text{tool}}$ covering tool invocations and $A_{\text{resp}}$ covering natural-language responses; $O$ is the observation space, comprising tool returns $r \in \mathcal{R}$ and user messages $u$; and $T(s_{t+1} \mid s_t, a_t)$ is the stochastic transition function, realized by a grounded executor $\varepsilon$. A trajectory $\tau = [(o_1, a_1), \ldots, (o_T, a_T)]$ records the full sequence of observation-action pairs. Because the agent cannot directly read the latent environment state $s$, it must ground arguments on $c_0$ (accessed indirectly through read-only tool returns or values relayed via user messages), prior tool-call returns $r$, or earlier user messages $u$. The synthesis goal is to produce a dataset $\mathcal{D} = \{(c_0^{(i)}, U^{(i)}, GT^{(i)})\}_i$, where $U^{(i)} = (u_k^{(i)})_k$ is the per-turn user-message sequence, such that every argument value $v$ carries an explicitly declared, executor-verifiable causal source.

\subsection{FSM Definition}
\label{sec:fsm}

While the POMDP above captures the full environment dynamics (tool-execution state, returns, user messages), the FSM abstracts over it at the dialogue-phase level: each FSM state represents a recognizable phase of the interaction rather than a concrete environment configuration. \cspg{} models this high-level dialogue structure as a \textbf{Finite State Machine (FSM)}:
\begin{equation}
\mathcal{F} = (\Sigma, \sigma_0, \Sigma_f, \Delta),
\end{equation}
where $\Sigma$ is a set of \emph{dialogue-phase} states, each instantiating one type of a predefined, domain-agnostic state-type taxonomy (Appendix~\ref{app:fsm-spec}), $\sigma_0 \in \Sigma$ the initial state, $\Sigma_f \subseteq \Sigma$ the set of accepting terminal states, and $\Delta$ the set of directed transition edges. The taxonomy is fixed across domains, while the concrete states and edges of $\mathcal{F}$ are generated per toolset at synthesis time.

\paragraph{States.} Each $\sigma \in \Sigma$ describes a \emph{phase} of the user--agent interaction, rather than a tool-level execution dependency. For instance, $\Sigma$ includes: $\sigma_{\text{init}}$ (session opened), $\sigma_{\text{logged\_in}}$ (user authenticated) and $\sigma_{\text{search\_done}}$ (search results retrieved).

\paragraph{Edges.} Each edge $\delta = (\sigma, \mathcal{C}_\delta, \sigma', \mathcal{T}_\delta) \in \Delta$ encodes one dialogue turn: $\mathcal{C}_\delta = [t_1, \dots, t_m]$ is the ordered tool list, and $\mathcal{T}_\delta$ is a Provenance Tag map that assigns to every argument a declared causal source from the type system of \S\ref{sec:provenance}. The first edge $\delta_0$ (from $\sigma_0$) may only declare \Pinit{} or \Pcreate{} sources, since no prior turns exist to reference.

\paragraph{Extensibility.} Unlike trace-first methods that rely on a hand-curated or signature-derived tool-dependency graph, the FSM here is constructed by $\mathcal{A}_{\text{FSM}}$ at synthesis time by matching each tool in the documentation to the predefined dialogue-phase state types it transitions between, without manually specifying pairwise tool dependencies; a new toolset simply triggers automatic rebuilding from its updated documentation.

\subsection{Provenance Tag Type System}
\label{sec:provenance}
For every argument $p$ in a tool call, its causal source satisfies:
\begin{multline}
\mathrm{src}(p) \in \\
\quad \{\Pinit,\; \Poutput(k', j', f'),\; \Pcreate,\;
\Pumsg(k''),\; \Pfallback(s_0)\}.
\end{multline}
Table~\ref{tab:argsource} summarizes the semantics. The four design-time types form $\mathcal{S}_{\text{decl}} = \{\Pinit,\; \Poutput,\; \Pcreate,\; \Pumsg\}$, which $\mathcal{A}_{\text{FSM}}$ declares for each turn; \Pfallback{} is excluded from $\mathcal{S}_{\text{decl}}$ and is produced only by $\mathcal{A}_{\text{plan}}$ at runtime when a declared source $s_0 \in \{\Pinit, \Poutput, \Pumsg\}$ turns out to be unusable. The planner then recovers the argument from another available initial-state value (\Pinit{}) or a newly introduced user-side value (\Pcreate{}) when permitted. The resulting call is executor-validated after binding. In all cases, the original declared source $s_0$ is preserved in \texttt{fallback\_from}. The \Pfallback{} rate probes how faithfully $\mathcal{A}_{\text{plan}}$ honors the declared sources; see App.~\ref{app:fallback-study}.

\begin{table}[t]
\centering\small
\setlength{\tabcolsep}{4pt}
\renewcommand{\arraystretch}{1.1}
\begin{tabular}{@{}c l p{4.4cm}@{}}
\toprule
\textbf{Sym.} & \textbf{Type} & \textbf{Semantics} \\
\midrule
\Pinit & initial\_state & A field of $c_0$; since the assistant cannot read $c_0$ directly, the value enters turn $k$ either from the user message or from a same-turn read-only call. \\
\Poutput & prev\_output & Field $f'$ of the $j'$-th call at a prior executed turn $k'\!<\!k$ (all of $k', j', f'$ are bound at runtime by $\mathcal{A}_{\text{plan}}$; $\mathcal{A}_{\text{FSM}}$ only declares the tag). \\
\Pcreate & self\_create & New user-side value introduced at this turn; later surfaced in $u_k$. \\
\Pumsg & prev\_user\_msg & Value semantically introduced in $u_{k''}$ for some $k''\!<\!k$; dialogue generated retroactively. \\
\Pfallback & fallback & Runtime fallback when a declared source $s_0 \in \{\Pinit, \Poutput, \Pumsg\}$ cannot be used: no matching field of $c_0$, no matching prior return, or no usable anchor value. The planner recovers the argument from another available \Pinit{} value or from a new \Pcreate{} value when permitted; the resulting call is executor-validated after binding, and the original source $s_0$ is kept in \texttt{fallback\_from}. \\
\bottomrule
\end{tabular}
\caption{Provenance Tag categories. \Pinit{} through \Pumsg{} are declarable by $\mathcal{A}_{\text{FSM}}$ at design time; \Pfallback{} is produced only by $\mathcal{A}_{\text{plan}}$ at runtime. Parametrized forms ($\Poutput(k',j',f')$, $\Pumsg(k'')$, $\Pfallback(s_0)$) are used only in the definition of \S\ref{sec:provenance}.}
\label{tab:argsource}
\end{table}

\subsection{Argument Dependency Graph}
Given a trajectory, we build a directed graph $\mathcal{G} = (V, E)$ over its tool-call arguments. Each node $v \in V$ is one argument-slot binding, a parameter consumed by some tool call at some turn. An edge $(u, v) \in E$ means that $v$'s bound value is reused from an earlier occurrence $u$ of the \emph{same} value instance, where $u$ is the upstream source of that value (which may be a field of a prior tool-call return, or a value the user first supplied in an earlier user message).

For each argument $v \in V$, let $\mathrm{turn}(v)$ be the turn in which $v$ is consumed and $\mathrm{origin}(v)$ be the turn in which the value carried by $v$ first entered the dialogue (via a user message or an earlier tool return). Initial-state values enter the dialogue in the turn that consumes them, so for them $\mathrm{origin}(v) = \mathrm{turn}(v)$. The chain length of $v$, the mean chain length, and the longest chain over the trajectory are
\begin{equation}
\begin{gathered}
L(v) = \mathrm{turn}(v) - \mathrm{origin}(v), \\
\bar{L} = \tfrac{1}{|V|}\sum_{v \in V} L(v), \quad L^\star = \max_{v \in V} L(v),
\end{gathered}
\end{equation}
i.e., $L(v)$ counts how many turn boundaries the value crosses before reaching its consumption point. For example, an argument consumed at turn~3 whose value was first introduced at turn~1 has $L=2$. We call $v$ a \emph{dependent argument} when $L(v) \ge 1$; in provenance-tag terms (\S\ref{sec:provenance}), arguments with \Pinit{}, \Pcreate{}, or \Pfallback{} sources always have $L(v)=0$ (their value is introduced in the consuming turn itself), while \Poutput{} and \Pumsg{} are the two sources of cross-turn dependencies. The fraction of dependent arguments in $V$ is a breadth-style measure of cross-turn grounding, complementary to chain length.

% =====================================================================
\section{Method}
\label{sec:method}

\subsection{Overview}
\label{sec:method-overview}

\cspg{} casts multi-turn tool-use data synthesis as a closed loop in which three specialized LLM agents cooperate with one executor (Figure~\ref{fig:pipeline}). $\mathcal{A}_{\text{FSM}}$ synthesizes a provenance-annotated FSM tool-call skeleton from the raw tool documentation. $\mathcal{A}_{\text{plan}}$ then fills, for each call, the executor-bound arguments $\boldsymbol{\theta}^{\text{exec}}$ and per-argument provenance metadata $\boldsymbol{\theta}^{\text{prov}}$, dispatches the call to $\varepsilon$, and groups same-turn calls into parallel step groups. After the call sequence is verified, $\mathcal{A}_{\text{msg}}$ synthesizes the trajectory's textual messages. The pipeline enforces a \textbf{Provenance Invariant}: before any argument value $v$ is committed, its causal source $\mathrm{src}(v)$ must be explicitly declared, and for the three upstream-grounded types (\Pinit, \Poutput, \Pumsg) the referenced upstream must exist or be pre-registered at synthesis time. The two ungrounded types (\Pcreate, \Pfallback) are still logged in $\boldsymbol{\theta}^{\text{prov}}$ so every value remains evaluable. This shifts cross-turn argument dependencies from a passively emerging side-effect into an explicitly planned property validated turn by turn.

\subsection{$\mathcal{A}_{\text{FSM}}$: FSM Skeleton Synthesis}
\label{sec:method-fsm}

$\mathcal{A}_{\text{FSM}}$ synthesizes the FSM $\mathcal{F}$ of \S\ref{sec:fsm} from the summarized tool documentation $\widetilde{\mathcal{T}}$ for the tool set $\mathcal{T}$ and the initial-state summary $\widetilde{c}_0$. For each edge $\delta$, it commits an ordered tool list $\mathcal{C}_\delta = [t_1, \dots, t_m]$ for the corresponding turn and a Provenance Tag map $\mathcal{T}_\delta$ assigning every required argument a tag from $\mathcal{S}_{\text{decl}}$ (Table~\ref{tab:argsource}). A lightweight verifier $\mathcal{V}_{\text{spec}}$ runs structural checks on the output and triggers regeneration (Appendix~\ref{app:vspec}).

\subsection{$\mathcal{A}_{\text{plan}} + \varepsilon$: Tool-Call Planning and Execution}
\label{sec:method-plan}

$\mathcal{A}_{\text{plan}}$ takes the skeleton $\mathcal{F}$ and a sampled path $\pi = (\delta_0, \dots, \delta_{N-1})$, obtained by a weighted random walk over $\mathcal{F}$ from $\sigma_0$ to some $\sigma_f \in \Sigma_f$ (transition weights and long-tail emphasis detailed in Appendices~\ref{app:fsm-spec},~\ref{app:vspec}), and converts it into an executable call sequence carried out by $\varepsilon$. For each turn, $\mathcal{A}_{\text{plan}}$ inspects the executed history $\mathcal{H}_{<k}$ and the declared tags, binds a value for each call, and groups intra-turn calls into parallel step groups (no intra-group \Poutput edges). $\varepsilon$ then executes group by group; on failure, only the offending call is regenerated and retried (Appendices~\ref{app:method-plan-fill} and \ref{app:method-plan-group}).

Two tag types require runtime binding. For \Poutput{}, $\mathcal{A}_{\text{FSM}}$ only declares that \emph{some} prior return is reused; $\mathcal{A}_{\text{plan}}$ binds the specific field at runtime against the actual return of $\varepsilon$. If no declared upstream satisfies the binding, it records \Pfallback{} and recovers the argument from another available \Pinit{} value or from a new \Pcreate{} value when permitted; a declared \Pinit{} source that cannot be located in $c_0$ falls back the same way. The executor validates the resulting call after the recovery value is bound (see App.~\ref{app:method-plan-fill}). This both surfaces the mismatch as a diagnostic signal in $\boldsymbol{\theta}^{\text{prov}}$ (\S\ref{sec:discussion}) and prevents a single binding failure from invalidating an entire trajectory. For \Pumsg{}, the value must appear in an earlier user message that $\mathcal{A}_{\text{msg}}$ has yet to generate; we break this cycle by fixing the value at planning time (via a temporary \Pcreate{} remap) and registering it together with a scheduled anchor turn $k^\star$ in a virtual history $\Delta_v$, which $\mathcal{A}_{\text{msg}}$ later renders into $u_{k^\star}$ (Appendix~\ref{app:method-plan-pum}).

\subsection{$\mathcal{A}_{\text{msg}}$: Post-hoc Synthesis of Dialogue Text Messages}
\label{sec:method-msg}

Once all $(t_i, \boldsymbol{\theta}^{\text{exec}}_i, r_i)$, $\boldsymbol{\theta}^{\text{prov}}_i$ and $\Delta_v$ are frozen, $\mathcal{A}_{\text{msg}}$ synthesizes the user message $u_k$ for every turn and an assistant natural-language response when $\mathcal{C}_k = \varnothing$. Because messages are written only after the ground truth is fixed, every message is grounded against actual returns. The user message $u_k$ is driven by a must-mention list derived from $\boldsymbol{\theta}^{\text{prov}}_i$ and $\Delta_v[k]$, with two regimes: \emph{newly introduced} values (\Pcreate, \Pfallback, the anchor side of a future \Pumsg{} consumer, and \Pinit{} values that must be passed verbatim to a tool) appear as explicit literals in $u_k$; \emph{referenced} values (\Poutput{} and the consumer side of \Pumsg{}) appear via vague referents whose unambiguity is verified by the validator. Full marker semantics and prompt-level constraints are in Appendix~\ref{app:method-msg}.

\subsection{Challenge Scenario Injection and Multi-Layer Validation}
\label{sec:method-challenge}

For each validated base trajectory, two rewrite operators synthesize hard variants targeting common deployment failure modes. $\Psi_{\text{param}}$ (MissParam) samples a few \Pcreate{} arguments from an originating turn $k^*$, replaces their literals in $u_{k^*}$ with vague references, and clears the turn-$k^*$ ground truth so that the assistant emits a clarification; an inserted reveal turn then re-supplies the missing values and re-binds the call. $\Psi_{\text{func}}$ (MissFunc) temporarily removes a tool that the base actually invokes and re-exposes it at the next turn via a user-side handoff message. Both operators are guarded by an executor replay that requires the rewritten outcome to match the original base (Appendix~\ref{app:method-challenge}).

Before being persisted, every trajectory passes through two complementary checks: a code-side deterministic check (legal provenance references, closed $\Delta_v$, minimum length), and an agent-side LLM judge checking intent alignment and rewrite-scenario structural compliance. Rejection by either check triggers immediate discard.

% =====================================================================
\section{Experiments}
\label{sec:exp}

\subsection{Setup}

\paragraph{Backbone and baselines.} We SFT on Qwen3-4B-Instruct-2507~\citep{yang2025qwen3}. The directly comparable open-source BFCL v4 multi-turn baselines at the 4B / 8B scale include AWM~\citep{2026awm}, CM2~\citep{2026cm2}, ToolACE-2 and BitAgent~\citep{2026fissiongrpo}, and TOUCAN~\citep{xu2025toucan}; we additionally report BFCL v3 multi-turn numbers for MAGNET~\citep{yin2025magnet}, ToolWeave~\citep{khandelwal2026toolweave}, and MUA-RL~\citep{2025muarl} as reference (listed in a separate row group in Table~\ref{tab:main}; not directly comparable due to benchmark version drift).

\paragraph{Synthesis pipeline.} The three agents are instantiated as $\mathcal{A}_{\text{FSM}} = $ Gemini-3.1 Pro~\citep{deepmind2026gemini31pro}, $\mathcal{A}_{\text{plan}} = $ Gemini-3 Pro~\citep{deepmind2025gemini3pro}, and $\mathcal{A}_{\text{msg}} = $ Qwen3-235B-A22B-Instruct-2507~\citep{yang2025qwen3}, chosen to balance FSM-source compliance and dialogue-style diversity (see App.~\ref{app:fallback-study}). Gemini-3.1 Pro attains the lowest fallback rate as the planner (2.02\% vs.\ 2.76\%, App.~\ref{app:fallback-study}), but its endpoint was less stable under high request concurrency; since a single FSM skeleton is amortized over several sampled trajectories while $\mathcal{A}_{\text{plan}}$ accounts for the bulk of the pipeline's request volume, we assign Gemini-3.1 Pro to the low-throughput FSM stage and Gemini-3 Pro to the high-throughput planning stage. The executor $\varepsilon$ wraps the original BFCL v4 multi-turn and $\tau^2$-bench backends without modification, so every synthesized trajectory is grounded against the same simulator used at evaluation. The initial configurations consumed at synthesis time are self-generated (following the released schemas) rather than taken from the benchmark test data, which is used only for evaluation. A single pipeline configuration is used across all source domains: only the tool documentation and the executor backend are swapped per domain, while $\mathcal{A}_{\text{FSM}}, \mathcal{A}_{\text{plan}}, \mathcal{A}_{\text{msg}}$, the FSM state-type taxonomy, and the Provenance Tag system remain identical. The released SFT mix contains $\approx 9$k trajectories.

\paragraph{Training.} We train with the verl framework~\citep{sheng2025hybridflow} on a single node equipped with eight 80GB GPUs. All experiments use full-parameter SFT with AdamW: learning rate $1\!\times\!10^{-6}$, global batch size $128$, and $10$ epochs.

\paragraph{Benchmarks and metrics.} BFCL v4 multi-turn (Base / MissFunc / MissParam / LongCtx) is scored with the official BFCL harness under the official environments and test data; we report the harness's multi-turn accuracy, which combines state-based checks on the executed environment with response-based checks on the emitted call sequence. $\tau^2$-bench Retail / Airline~\citep{barres2025tau2bench} is evaluated under the official harness with the think tool disabled; we report pass$^1$ only. The argument-value chain-length diagnostics $\bar{L}$ and $L^\star$ (\S\ref{sec:prelim}) are reported as a synthesis-side measurement in Fig.~\ref{fig:ppg-comparison}.

\subsection{Main Results}
\label{sec:main}

\begin{table*}[!tbp]
\centering\scriptsize
\setlength{\tabcolsep}{3.5pt}
\renewcommand{\arraystretch}{1.0}
\resizebox{0.90\textwidth}{!}{%
\begin{tabular}{l c ccccc | ccc}
\toprule
& & \multicolumn{5}{c}{\textbf{BFCL v4 Multi-Turn}} & \multicolumn{3}{c}{\textbf{$\tau^2$-bench (pass$^1$)}} \\
\cmidrule(lr){3-7}\cmidrule(lr){8-10}
\textbf{Model} & \textbf{Size} & \textbf{Avg} & \textbf{Base} & \textbf{MFn} & \textbf{MPm} & \textbf{LCtx} & \textbf{Retail} & \textbf{Airline} & \textbf{Avg} \\
\midrule
\multicolumn{10}{l}{\emph{Closed-source / large-scale reference (BFCL v4)}} \\
GPT-5.2-High          & --   & 48.5 & -- & -- & -- & -- & 81.6 & 62.5 & 72.1 \\
Claude Sonnet 4.5           & --   & 61.4 & 69.0 & 65.0 & 52.5 & 59.0 & 86.2 & 70.1 & 78.2 \\
Gemini-3 Pro                & --   & 60.8 & 64.5 & 60.0 & 54.5 & 64.0 & 85.3 & 72.7 & 79.0 \\
DeepSeek-V3.2-Exp           & 685B & 37.4 & 41.5 & 39.5 & 33.5 & 35.0 & --   & --   & --   \\
Qwen3-235B-A22B             & 235B & 45.4 & --   & --   & --   & --   & 71.9 & 45.6 & 58.8 \\
\midrule
\multicolumn{10}{l}{\emph{BFCL v3 MT (reference)}} \\
ToolWeave-8B                & 8B  & 21.1 & 28.0 & 18.0 & 18.5 & 20.0 & --   & --   & --   \\
MAGNET-7B-SFT               & 7B  & 26.5 & 35.5 & 24.0 & 27.5 & 19.0 & --   & --   & --   \\
MAGNET-7B-mDPO              & 7B  & 27.8 & 39.0 & 24.0 & 26.0 & 22.0 & --   & --   & --   \\
MAGNET-14B-SFT              & 14B & 33.4 & 47.0 & 32.0 & 32.0 & 22.5 & --   & --   & --   \\
MAGNET-14B-mDPO             & 14B & 37.9 & 52.0 & 36.0 & 35.5 & 28.0 & --   & --   & --   \\
MUA-RL-8B                   & 8B  & 14.6 & 21.0 & 11.5 & 15.0 & 11.0 & 49.8 & 19.0 & 34.4 \\
MUA-RL-14B                  & 14B & 25.3 & 40.5 & 14.0 & 25.0 & 21.5 & 66.0 & 38.0 & 52.0 \\
MUA-RL-32B                  & 32B & 28.4 & 42.0 & 20.0 & 30.0 & 21.5 & 67.3 & 45.4 & 56.4 \\
\midrule
\multicolumn{10}{l}{\emph{BFCL v4 MT (directly comparable)}} \\
CM2-8B-RL                   & 8B  & 36.5 & 44.5 & 32.0 & 35.0 & 34.5 & 36.4 & 27.0 & 31.7 \\
CM2-8B-SFT                  & 8B  & 26.8 & 30.0 & 27.5 & 24.5 & 25.0 & 19.5 & 23.5 & 21.5 \\
ToolACE-2-8B                & 8B  & 37.0 & 47.0 & 31.0 & 28.0 & 42.0 & 8.2  & 26.7 & 17.5 \\
BitAgent-8B                 & 8B  & 37.8 & 46.5 & 37.5 & 24.0 & 43.0 & 6.1  & 37.3 & 21.7 \\
TOUCAN-7B                   & 7B  & 24.0 & 30.0 & 21.0 & 24.0 & 21.0 & 22.8 & 20.0 & 21.4 \\
AWM-4B                      & 4B  & 30.3 & 37.5 & 35.5 & 28.5 & 19.5 & 30.3 & 19.0 & 24.7 \\
AWM-8B                      & 8B  & 40.1 & 49.5 & 48.0 & 36.0 & 27.0 & 41.2 & 33.5 & 37.4 \\
\midrule
Qwen3-4B-Instruct-2507    & 4B   & 22.1 & 26.5 & 21.0 & 15.5 & 25.5 & 40.4 & 24.0 & 32.2 \\
\rowcolor{gray!12}\textbf{\cspg-4B SFT (ours)} & 4B & \textbf{30.4} & \textbf{38.0} & \textbf{23.0} & \textbf{24.5} & \textbf{36.0} & \textbf{42.1} & \textbf{28.0} & \textbf{35.1} \\
\textit{$\Delta$ (Ours $-$ backbone)}  & & {\color{ForestGreen}+8.3} & {\color{ForestGreen}+11.5} & {\color{ForestGreen}+2.0} & {\color{ForestGreen}+9.0} & {\color{ForestGreen}+10.5} & {\color{ForestGreen}+1.7} & {\color{ForestGreen}+4.0} & {\color{ForestGreen}+2.9} \\
\bottomrule
\end{tabular}%
}
\caption{Main results on BFCL v4 Multi-Turn and $\tau^2$-bench. MFn, MPm, and LCtx denote MissFunc, MissParam, and LongCtx, respectively; Retail and Airline are the two $\tau^2$-bench domains. Row groups separate BFCL v4 (directly comparable) from BFCL v3 (reference). \textbf{Sources:} GPT-5.2~\citep{openai2025gpt52card}, Claude Sonnet 4.5~\citep{anthropic2025claude45}, Gemini-3 Pro~\citep{deepmind2025gemini3pro}, DeepSeek-V3.2-Exp~\citep{deepseekai2025deepseekv32pushingfrontieropen}, Qwen3-235B-A22B and backbone~\citep{yang2025qwen3}; ToolWeave~\citep{khandelwal2026toolweave}, MAGNET~\citep{yin2025magnet}, MUA-RL~\citep{2025muarl}; ToolACE-2 / BitAgent results from~\citep{2026fissiongrpo}, CM2~\citep{2026cm2}, TOUCAN~\citep{xu2025toucan}, AWM~\citep{2026awm}.}
\label{tab:main}
\end{table*}
Over the Qwen3-4B-Instruct backbone, \cspg-4B SFT improves BFCL v4 multi-turn Avg by +8.3 (22.1 $\to$ 30.4) and $\tau^2$-bench Avg by +2.9 (32.2 $\to$ 35.1) without any RL. On BFCL v4 the gains concentrate on Base (+11.5), LongCtx (+10.5), and MissParam (+9.0), while MissFunc shows a smaller lift (+2.0). The $\tau^2$-bench improvement (Retail +1.7 / Airline +4.0) is more modest, consistent with the benchmark's long-horizon state-tracking demands; \cspg-4B's competitive $\tau^2$ result under a 4B SFT-only setup indicates that argument-level provenance provides a meaningful signal even at this small scale. As a complementary out-of-distribution probe, we evaluate \cspg-4B on the BFCL v4 single-turn track (Appendix~\ref{app:ood-singleturn}): multi-turn-only training does not regress single-turn performance and yields a small consistent lift on both Non-live and Live splits.

Among methods evaluated on the same BFCL v4 MT release, \cspg-4B is on par with same-parameter-scale baselines on BFCL v4 MT Avg (30.4 vs.\ AWM-4B 30.3) while substantially outperforming them on $\tau^2$-bench (35.1 vs.\ AWM-4B 24.7), where long-horizon state-tracking is the bottleneck. \cspg-4B still trails 8B baselines on BFCL v4, but on $\tau^2$ it matches or exceeds most of them (CM2-8B-RL 31.7, ToolACE-2-8B 17.5, BitAgent-8B 21.7) despite using only SFT on a 4B backbone with 9k trajectories. We attribute the BFCL v4 gap primarily to scale rather than data quality.

The closed-source frontier (Claude Sonnet 4.5, Gemini-3 Pro) reaches BFCL v4 MT Avg in the 60s and $\tau^2$ Avg in the 70s, indicating that the open-source 4B regime still has substantial headroom on BFCL v4 multi-turn. Notably, GPT-5.2-High shows a wider gap to the Claude/Gemini frontier on BFCL v4 MT (48.5 vs.\ $\sim$60) than on $\tau^2$ (72.1 vs.\ $\sim$78), suggesting the two benchmarks exercise partly orthogonal capabilities.

\subsection{Ablations}
\label{sec:ablation}

\begin{table}[!t]
\centering\footnotesize
\setlength{\tabcolsep}{3pt}
\begin{tabular}{l ccccc}
\toprule
\textbf{Variant} & \textbf{Avg} & \textbf{Base} & \textbf{MFn} & \textbf{MPm} & \textbf{LCtx} \\
\midrule
\rowcolor{gray!12}\textbf{\cspg-4B (Full)}     & \textbf{30.4} & \textbf{38.0} & \textbf{23.0} & \textbf{24.5} & \textbf{36.0} \\
\;\;A1: w/o Prov.\ Tag    & 24.0 & 29.5 & 19.5 & 21.5 &  25.5\\
\;\;A2: w/o $\Psi$        & 28.8 & 37.0 & 21.0 & 22.5 &  34.5\\
\bottomrule
\end{tabular}
\caption{Ablation on BFCL v4 Multi-Turn at matched data scale. A1 removes per-argument Provenance Tag declarations; A2 removes both rewrite operators $\Psi_{\text{param}},\Psi_{\text{func}}$.}
\label{tab:ablation}
\end{table}

\paragraph{Note on $\mathcal{A}_{\text{FSM}}$ and pipeline-level ablations.}
We do not include an $\mathcal{A}_{\text{FSM}}$ ablation row in Table~\ref{tab:ablation}: removing the FSM skeleton (i.e., reverting to free-form turn-by-turn generation by $\mathcal{A}_{\text{plan}}$) cuts the pipeline's data pass-rate from $89\%$ to $50\%$ (see Appendix~\ref{app:pipeline-ablation}), and the surviving trajectories are too few and too biased to support a fair SFT comparison. Similarly, disabling per-call retry (the ``noRetry'' variant) drops the pass-rate to $54\%$. Removing Provenance Tags while keeping the FSM (``FSM+noTag'') yields a high pass-rate ($92\%$) but produces zero \Poutput{} arguments, confirming that the tag declarations are the primary mechanism driving cross-turn dependency generation. We treat $\mathcal{A}_{\text{FSM}}$ as a non-optional structural component and report only the two downstream ablations below.

\paragraph{A1: Provenance Tag declarations.}
Removing the per-argument Provenance Tag declarations that $\mathcal{A}_{\text{FSM}}$ commits in $\mathcal{T}_\delta$ (\S\ref{sec:method-fsm}) and letting $\mathcal{A}_{\text{plan}}$ choose argument sources freely causes a substantial drop across all four subsets (Avg $30.4 \to 24.0$, $-6.4$). The hit is sharpest on LongCtx ($-10.5$) and Base ($-8.5$), with MissFunc ($-3.5$) and MissParam ($-3.0$) showing moderate declines, confirming that the tag declarations act as a structural prior that constrains $\mathcal{A}_{\text{plan}}$ to ground each argument in a verifiable upstream, most useful precisely where long-horizon argument tracking matters. App.~\ref{app:fallback-study} reports a complementary backbone-level study where the \Pfallback{} rate correlates with these declaration-compliance failures.

\paragraph{A2: Rewrite operators $\Psi_{\text{param}}, \Psi_{\text{func}}$.}
Removing the rewrite operators $\Psi_{\text{param}}$ and $\Psi_{\text{func}}$ (\S\ref{sec:method-challenge}) at matched data scale lowers Avg to $28.8$ ($-1.6$). The effect is modest but consistent across MissFunc ($-2.0$), MissParam ($-2.0$), Base ($-1.0$), and LongCtx ($-1.5$), matching the expectation that the operators specifically target deployment failure modes while leaving standard multi-turn performance largely intact. The smaller magnitude relative to A1 also indicates that the bulk of the gain comes from the structural Provenance Tag constraint, not from the hard-scenario rewrites alone.

\subsection{Case Study}
\label{sec:case-study}

\begin{figure*}[t]
  \centering
  \includegraphics[width=1.0\linewidth]{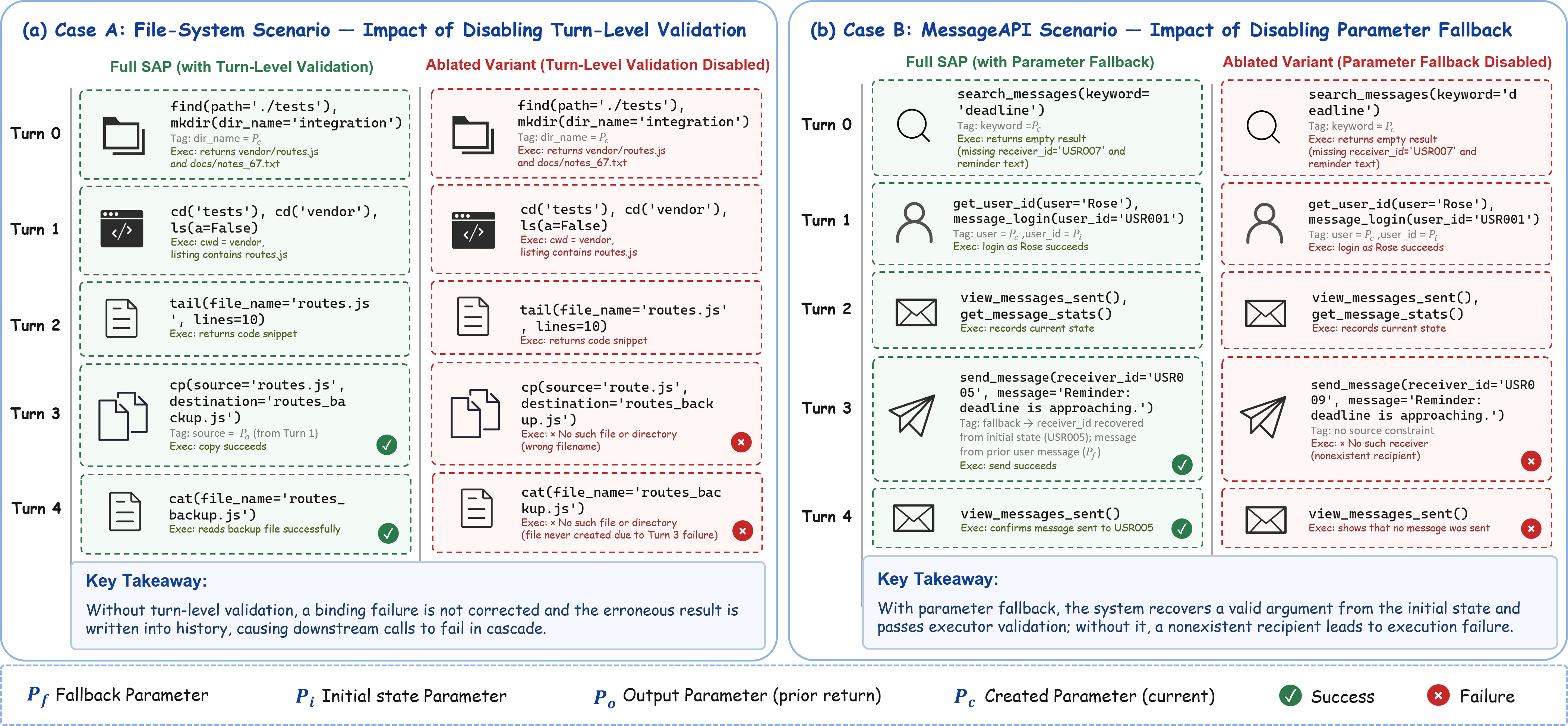}
  \caption{Case study of \cspg{}'s error localization and hybrid validation. (a) Turn-level execution validation prevents an incorrect file binding from cascading to a later call. (b) Parameter fallback recovers a valid recipient and is checked by the executor, whereas disabling it causes message delivery to fail.}
  \label{fig:case}
\end{figure*}

Figure~\ref{fig:case} provides qualitative evidence for \cspg{}'s provenance constraints and execution-in-the-loop validation. In (a), full \cspg{} resolves the \texttt{cp} source \texttt{routes.js} as \Poutput{} from the earlier \texttt{ls} return; validation confirms the copy, so the backup can be read on the next turn. Without turn-level validation, a one-character error (\texttt{route.js}) causes the copy and the subsequent \texttt{cat} call to fail. In (b), when the initial search provides no usable recipient, \cspg{} records \Pfallback{}, recovers \texttt{USR005} from the initial state, and validates the send successfully. Disabling fallback leaves \texttt{USR009} ungrounded and nonexistent, causing both message delivery and subsequent verification to fail.

% =====================================================================
\section{Discussion}
\label{sec:discussion}

\cspg{} does not fully prevent argument hallucinations at newly created values (\Pcreate{}) or at fallback recoveries that must introduce a new value. When a declared upstream is unavailable, \Pfallback{} instead records the recovery event and its original source, whether the recovered binding selects another available value or introduces a new one; the resulting call is then checked by the executor. This localizes unresolved argument quality to a single call and makes every recovery auditable in $\boldsymbol{\theta}^{\text{prov}}$. Hallucinations cannot propagate along \Poutput{} (blocked by real execution) or \Pumsg{} (blocked by the \S\ref{sec:method-msg} consistency check). This contrasts with committee-style validators that reduce hallucination probability without bounding its propagation scope.

We train on $\approx$9k trajectories with SFT only, no RL, a deliberately compact setup imposed by the per-trajectory live-executor cost (\S\ref{sec:limitations}) and our compute budget. Under this small-model, small-data, SFT-only setting, \cspg-4B still matches or exceeds several 8B RL-trained methods on $\tau^2$-bench, particularly where long-horizon state-tracking aligns with our provenance design. That this holds \emph{without} scale or RL confounders gives a clean lower-bound estimate of the method's intrinsic data-quality contribution.

Most open-source 4B baselines (e.g., AWM-4B) are think-mode RL-trained models with long deliberation chains before each tool call. \cspg-4B trains from a non-think backbone (Qwen3-4B-Instruct-2507) with SFT only, so its per-turn inference cost is substantially lower at comparable or better $\tau^2$-bench accuracy.

% =====================================================================
\section{Conclusion}
\label{sec:conclusion}
We presented \cspg{}, a state-guided pipeline for multi-turn tool-use trajectory synthesis that promotes argument provenance from a post-hoc annotation to a synthesis-time active constraint. In addition, \cspg{} is domain-agnostic: rebuilding the FSM from updated tool documentation and swapping the executor backend is sufficient to support new toolsets, with no manual rule engineering. The mechanism rests on two components: (i)~the Provenance Tag type system, which explicitly declares the causal source of every argument; and (ii)~FSM-driven structure-first generation, which dynamically builds a provenance-annotated skeleton from tool documentation and enforces per-turn executability via execution-in-the-loop validation. Experiments and ablations on BFCL v4 multi-turn and $\tau^2$-bench validate the design.

The framework opens two directions. \emph{(i)~\cspg{} as a compiler for agentic RL.} Provenance Tag serves as a process-reward interface: successful \Poutput{} retrievals are positive signals, and argument-binding mismatches localize provenance errors, supporting an SFT cold-start $\to$ RL with process rewards curriculum without extra annotation, since the signals are written into $\boldsymbol{\theta}^{\text{prov}}$ at synthesis time. \emph{(ii)~Scaling and pipeline extensions.} Data scaling, larger backbones, RL, \emph{execution-conditioned FSMs} that revise the skeleton when turns fail, and \emph{depth-guided synthesis} that uses the dependency-graph diagnostic as an optimization objective.

% =====================================================================
\section*{Limitations}
\label{sec:limitations}

We train \cspg{} only on a 4B backbone; the 8B-and-larger numbers in Table~\ref{tab:main} are quoted from the original papers, as most concurrent data-synthesis works (FunReason-MT, MAGNET, ToolWeave) release no full pipeline. Methodologically, $\mathcal{A}_{\text{FSM}}$ drafts the skeleton in one pass, so dependencies that surface only after partial execution (e.g., return-code-conditioned branches) are downgraded to \Pfallback{} at runtime instead of a grounded \Poutput{} edge; Provenance Tag covers tool-call arguments only, not entity anaphora in free text; and all runs use English documentation in single-agent settings.

Following standard practice, the data-synthesis pipeline reuses the BFCL and $\tau^2$-bench schemas and executor backends as its environment, though initial states, user messages, and call sequences are sampled independently rather than copied from the test split. Two costs remain: live-executor validation is expensive in wall-clock and API terms (App.~\ref{app:cost-study}), and we do not report human-agreement statistics for the single LLM judge of \S\ref{sec:method-challenge}.
% =====================================================================
\section*{AI Use Statement}
\label{sec:ai-use}
In line with the ACL Policy on AI Writing Assistance, large language models were used as writing aids for grammar polishing, paragraph restructuring, and literature search, and for code completion when implementing the synthesis pipeline. All experimental design, methodological contributions, hypothesis formulations, error analyses, and final claims are the authors' work; no content was generated end-to-end by AI without subsequent verification and editing.

% =====================================================================
% \section*{Acknowledgments}
% =====================================================================
\bibliography{acl_latex}

\appendix

\section*{Appendix}

\section{\cspg{} Pipeline Details}
\label{app:method-details}

This appendix expands the design choices behind the three agents and the rewrite operators of \S\ref{sec:method}.

\subsection{Detailed FSM Specification}
\label{app:fsm-spec}

The FSM of \S\ref{sec:fsm} is realized by mapping each domain's tool documentation to a shared, domain-agnostic \emph{state-type taxonomy} that lists which dialogue phases may appear in any tool-use scenario. New domains reuse the same taxonomy and downstream protocol; supplying updated tool documentation is sufficient to regenerate the corresponding states and transitions, without manually rewriting rule code.

\paragraph{State-type taxonomy ($\Sigma$).} We use 13 abstract types reused across all domains: \texttt{INITIAL} (session start); \texttt{AUTH\_REQUIRED} / \texttt{AUTH\_COMPLETED} (authentication phases); \texttt{INFO\_GATHERING} (waiting for user-supplied arguments); \texttt{SEARCHING} (read-only lookup in progress); \texttt{FOUND} / \texttt{UNAVAILABLE} (lookup outcomes); \texttt{ACTION\_REQUIRED} (irreversible action pending user confirmation); \texttt{ACTION\_COMPLETED} (action persisted by the executor); \texttt{ERROR} (recoverable executor error); \texttt{REJECTION} (out-of-scope or invalid request); \texttt{COMPLETED} (terminal success); and \texttt{NORMAL} (catch-all for domain-specific intermediate phases). Each $\sigma \in \Sigma$ instantiates exactly one of these types.

\paragraph{Transition record ($\delta \in \Delta$).} Each transition carries six fields beyond the source/target states:
\begin{itemize}[leftmargin=*,topsep=2pt,itemsep=1pt]
\item \texttt{action}: the tool (or composite tool list $\mathcal{C}_\delta$) whose invocation realizes the transition.
\item \texttt{condition}: a natural-language guard ($\mathcal{A}_{\text{plan}}$ uses this in-context).
\item \texttt{probability}: base sampling probability used by the weighted random walk over $\mathcal{F}$.
\item \texttt{weight}: long-tail emphasis weight (see \S\ref{app:vspec}).
\item \texttt{is\_critical}: marks rare but high-value paths (e.g., authentication failure, payment-error branches) that the long-tail emphasis subset \(\mathcal{E}\) preferentially covers.
\item \texttt{provenance\_tag}: realizes $\mathcal{T}_\delta$ at the engineering level, encoded as a nested map \texttt{\{tool:\{arg: tag\}\}} where \texttt{tag $\in$ \{initial\_state, prev\_output, self\_create, prev\_user\_msg\}} (the four design-time entries of $\mathcal{S}_{\text{decl}}$; \Pfallback{} is reserved for runtime and never appears here).
\end{itemize}

\paragraph{Cross-domain reuse.} Our implementation provides concrete FSMs for the $\tau^2$-bench Retail and Airline domains and for the BFCL v4 multi-turn domains (e.g., Twitter, file system), all instantiating the taxonomy above. The Retail FSM, for example, defines $|\Sigma|\!=\!25$ states (including domain-specific instances such as \texttt{auth\_email\_prompt}, \texttt{info\_gathering\_order\_id}, \texttt{search\_order}, \texttt{order\_found}, \texttt{action\_confirm\_cancel}, \texttt{cancel\_success}) and $|\Delta|\!=\!35$ transitions; the Airline and BFCL multi-turn FSMs reach comparable scale. A representative transition record:
\begin{Verbatim}[fontsize=\scriptsize]
{
  "from_state": "info_gathering_order_id",
  "to_state":   "search_order",
  "action":     "search_order",
  "condition":  "user supplies order ID",
  "probability": 1.0, "weight": 1.0,
  "is_critical": false,
  "provenance_tag": {
    "search_order": {
      "order_id": "prev_user_msg"
    }
  }
}
\end{Verbatim}
Adding a new domain therefore requires only supplying its tool documentation for this mapping; no change to $\mathcal{S}_{\text{decl}}$, the executor protocol, or downstream agents is needed.

\subsection{$\mathcal{V}_{\text{spec}}$: Structural Constraints and Long-Tail Coverage}
\label{app:vspec}

The verifier $\mathcal{V}_{\text{spec}}$ rejects an FSM $\mathcal{F}$ produced by $\mathcal{A}_{\text{FSM}}$ on three routine sanity checks (topological legality of $\Sigma, \Delta$; existence of at least one length-$N$ path from $\sigma_0$; argument completeness of $\mathcal{T}_\delta$ for every required argument) and triggers regeneration. A more substantive constraint forbids \Poutput{} and \Pumsg{} on the first edge (outgoing from $\sigma_0$). This is the most frequent failure mode we observed in $\mathcal{A}_{\text{FSM}}$'s raw output: because the FSM is drafted in one pass, the very first turn is often declared to reuse a prior return or an earlier user message even though no prior turn exists. Rejecting such skeletons at the verifier removes this ``first-turn reference to non-existent context'' error before any planning cost is incurred.

\paragraph{Long-tail coverage via inverse-frequency emphasis.} Pure unconstrained sampling lets head tools dominate $\mathcal{F}$ and starves rare tools of training signal. We maintain a flat call-frequency histogram $H : \mathcal{T} \to \mathbb{N}$ across all synthesized trajectories and, for each new FSM, sample an emphasis subset $\mathcal{E} \subseteq \mathcal{T}$ under inverse-frequency weighting; $\mathcal{A}_{\text{FSM}}$ must commit at least one turn whose call list intersects $\mathcal{E}$. $H$ encodes only tool frequencies, not inter-tool dependencies, so the resulting bias does not leak any graph structure back into $\mathcal{F}$.

\subsection{$\mathcal{A}_{\text{plan}} + \varepsilon$: Filling Protocol}
\label{app:method-plan-fill}

$\mathcal{A}_{\text{plan}}$ fills each call $t_i$ in turn $\delta_k$ in a refill-on-failure loop. Two design choices are worth highlighting.

\paragraph{Two parallel outputs.} The planner emits two parallel objects per call:
\begin{equation}
(\boldsymbol{\theta}^{\text{exec}}_i, \boldsymbol{\theta}^{\text{prov}}_i) \sim \mathcal{A}_{\text{plan}}\!\left(t_i, \mathcal{T}_\delta[t_i], \mathcal{H}_{<k}, c_0, \ell_{\text{last}}\right),
\end{equation}
where $\boldsymbol{\theta}^{\text{exec}}_i$ holds the executor-bound argument values and $\boldsymbol{\theta}^{\text{prov}}_i$ holds the per-argument Provenance Tag and (if applicable) the upstream reference. Decoupling the two tracks lets the executor proceed with $\boldsymbol{\theta}^{\text{exec}}_i$ while every fallback escalation or upstream rebinding is logged into $\boldsymbol{\theta}^{\text{prov}}_i$ without disturbing dispatch.

\paragraph{Error feedback as in-context signal.} On a failed call, the executor error $\ell_{\text{last}}$ is fed back into the next prompt rather than discarded. This converts the executor into an in-context critic for the planner LLM, which empirically removes most repeated schema/type errors within a couple of refills. If a call still fails after the cap, the trajectory is truncated at the current turn, and trajectories left too short are dropped to avoid degenerate samples.

\subsection{$\mathcal{A}_{\text{plan}}$: Intra-Turn Parallelization Grouping}
\label{app:method-plan-group}

Many tool-set backends admit multiple parallel calls within a single turn (e.g., independent lookups). After all calls of a turn are bound (i.e., arguments filled and validated), $\mathcal{A}_{\text{plan}}$ partitions them into ordered groups $(G_1, \dots, G_p)$ such that no \Poutput{} edge lies within a group (at execution time, intra-group returns become visible only after the entire group completes), with groups ordered by minimum call index.

\subsection{$\mathcal{A}_{\text{plan}}$: Resolving the \Pumsg{} Dependency Cycle}
\label{app:method-plan-pum}

The \Pumsg{} type is the only source whose dialogue rendering lies outside $\mathcal{A}_{\text{plan}}$'s purview, since user messages are synthesized later by $\mathcal{A}_{\text{msg}}$. We resolve the cycle through a remap-bind-restore protocol mediated by a virtual history $\Delta_v$.

\paragraph{Virtual history $\Delta_v$.} $\Delta_v$ is a per-turn list of \emph{pending introductions}: $\Delta_v[k] = \{(p_1, v_1), (p_2, v_2), \dots\}$, meaning ``$u_k$ must explicitly introduce values $v_1, v_2, \dots$ for downstream consumption.'' $\mathcal{A}_{\text{plan}}$ populates $\Delta_v$ during planning; $\mathcal{A}_{\text{msg}}$ later consumes it in turn order.

\paragraph{Remap-bind-restore.} For each argument $p$ with $\mathrm{src}(p) = \Pumsg$ at turn $k$, the planner first samples an anchor turn $k^\star < k$, avoiding turns that already anchor an unrelated value of the same surface form (otherwise the later disambiguator could not tell two homonymous anchors apart). It then \emph{temporarily} sets $\mathrm{src}(p) \leftarrow \Pcreate$ and invokes the filling protocol of Appendix~\ref{app:method-plan-fill} to obtain a concrete value $v$ that passes executor validation. Finally, it restores the tag to \Pumsg{} (recording $k^\star$ as the anchor turn) and appends $(p, v)$ to $\Delta_v[k^\star]$.

\paragraph{Closure check.} After all turns are planned, a closure check verifies that every \Pumsg{} argument with anchor $k^\star$ has a matching pair in $\Delta_v[k^\star]$ and that $\mathcal{A}_{\text{msg}}$ later emits $v$ as an explicit literal in $u_{k^\star}$ under normalized (case- and whitespace-insensitive) matching. Closure failures roll back the trajectory. Conceptually, the trick is that \Pcreate{} depends on no upstream, so reassigning the tag yields a value that is both executor-valid and free to appear anywhere earlier in the dialogue; $\Delta_v$ is the bridge that turns this freedom into a contract honored by $\mathcal{A}_{\text{msg}}$.

\subsection{$\mathcal{A}_{\text{msg}}$: Must-Mention Dispatch and Per-Call Synthesis}
\label{app:method-msg}

$\mathcal{A}_{\text{msg}}$ synthesizes $u_k$ under a per-turn must-mention list $\mathcal{M}_k$ derived from $\boldsymbol{\theta}^{\text{prov}}$ and $\Delta_v[k]$. Four markers cover all surface behaviors of values in $u_k$ (Table~\ref{tab:mustmention}). \Pinit values that the assistant can discover via auxiliary read-only calls (e.g., listing the current directory) are dropped from $\mathcal{M}_k$; only those that must be passed verbatim into a tool argument (IDs, credentials, token-like fields) are marked \texttt{[FROM\_CONFIG]}.

\begin{table*}[t]
\centering\footnotesize
\begin{tabular}{p{2.8cm}p{5.2cm}p{5.2cm}}
\toprule
\textbf{Marker} & \textbf{Trigger (when processing turn $k$)} & \textbf{Expression constraint in $u_k$} \\
\midrule
\texttt{[NEW]} & $\mathrm{src}(p) = \Pcreate$, \emph{or} $(p, v) \in \Delta_v[k]$ (anchor for downstream \Pumsg) & Explicit literal value: first occurrence in the dialogue. \\
\texttt{[FROM\_HISTORY]} & $\mathrm{src}(p) \in \{\Poutput,\, \Pumsg\}$ (upstream at turn $<k$) & Vague reference (``that file'', ``the ID you found earlier''); literal must NOT be repeated. \\
\texttt{[FROM\_CONFIG]} & $\mathrm{src}(p) = \Pinit$, value must be passed verbatim to a tool argument & Explicit literal value (the assistant cannot read $c_0$): IDs, credentials, token-like fields. \\
\texttt{[FALLBACK\_from\_$s_0$]} & $\mathrm{src}(p) = \Pfallback$ (original source $s_0$ in metadata) & Treated as \texttt{[NEW]}: explicit literal value. \\
\bottomrule
\end{tabular}
\caption{Must-mention markers dispatched by the orchestrator to $\mathcal{A}_{\text{msg}}$ for user-message synthesis.}
\label{tab:mustmention}
\end{table*}

\paragraph{Per-call decomposition.} A naive prompt would hand $\mathcal{A}_{\text{msg}}$ the entire $\mathcal{M}_k$ and ask for $u_k$ in one shot; in practice this causes the LLM to overload the front of $u_k$ with literals and skip later constraints. We instead slice $\mathcal{M}_k$ per call,
\begin{equation}
\mathcal{M}_k = \bigsqcup_{i=1}^{n} \mathcal{M}_k^{(i)},
\end{equation}
where $\mathcal{M}_k^{(i)}$ contains only the slots that $c_i$ actually consumes, generate an intent fragment $\phi_i$ for each $c_i$ under its slice, and concatenate the fragments with sequencing connectives:
\begin{equation}
u_k = \mathrm{Compose}(\phi_1, \dots, \phi_n),
\end{equation}
\begin{equation}
\phi_i = \mathcal{A}_{\text{msg}}(c_i \mid \mathcal{H}_{<k}, \mathcal{M}_k^{(i)}).
\end{equation}
The per-call decomposition yields by construction the property that every ground-truth call has a corresponding semantic expression in $u_k$, which removes the need for an additional ``call coverage'' post-check.

\paragraph{Four prompt-level constraints.} On top of $\mathcal{M}_k^{(i)}$, the fragment prompt enforces four semantic constraints that empirically suppress the most common degenerations: (i)~$u_k$ is the sole signal driving the assistant on tool-call turns, so it cannot rely on assistant clarification; (ii)~every entry in $\mathcal{M}_k^{(i)}$ must be matched by an expression in $\phi_i$; (iii)~the semantic direction of $\phi_i$ must agree with $c_i$'s actual operation (when $c_i$ is \texttt{unfollow}, $\phi_i$ must not express ``follow''); (iv)~when $u_k$ belongs to a different intent family from prior turns (e.g., search$\to$delete), the shift must be expressed explicitly. These collectively prevent silently mis-aligned user messages from passing the downstream LLM judge.

\paragraph{Assistant text responses.} On turns with $\mathcal{C}_k = \varnothing$, $\mathcal{A}_{\text{msg}}$ uses a dedicated prompt (e.g., \texttt{ASSIST\_CLARIFY\_PROMPT} for MissParam) and is required to (i)~address the immediately preceding user intent (e.g., request the missing argument), (ii)~introduce no new tool call or capability promise, and (iii)~not disclose information that the real tool returns have not exposed. The prior context visible to $\mathcal{A}_{\text{msg}}$ is the real executed history (calls, arguments, returns, descriptions), so both user and assistant text are grounded against actual returns rather than speculative ones.

\subsection{Challenge-Scenario Rewrite Operators}
\label{app:method-challenge}

Both rewrite operators take a validated base trajectory and produce a hard variant; the salient trick they share is an executor-replay check: the rewritten trajectory is re-executed through $\varepsilon$ and accepted only when its execution outcome matches the original base. Without this check, vague-reference rewrites or toolkit removals can silently produce unsolvable trajectories.

\paragraph{$\Psi_{\text{param}}$ (MissParam).} The operator samples one or more arguments with $\mathrm{src}(p) = \Pcreate$ from the base, replaces the corresponding literals in $u_{k^*}$ with vague references (``\texttt{report.pdf}'' $\to$ ``that file''), and empties turn $k^*$'s ground-truth call list. The assistant therefore emits a clarification request synthesized by $\mathcal{A}_{\text{msg}}$ in place of a tool call. A new reveal turn is inserted at position $k^* + 1$ in which $\mathcal{A}_{\text{msg}}$ generates a natural user message surfacing the missing values, and the original call is rebound at this turn. Multiple missing arguments may be revealed in a single turn or spread across several reveal turns; the operator is agnostic to the dataset's exact clarification format (free text, a \texttt{request\_for\_info} slot, etc.).

\paragraph{$\Psi_{\text{func}}$ (MissFunc).} The operator removes a tool $t^*$ that the base actually invokes from the visible toolkit at turn $k^*$, so the assistant cannot proceed. A user-side handoff message is injected whose semantic event is ``new tools have been registered into the visible toolkit,'' re-exposing $\mathcal{T}$; the assistant then invokes $t^*$ at $k^* + 1$. Both the validator and the training objective are driven by this semantic event, not the surface wording, so alternative carriers (a system-side marker, an external tool-registry event) are interchangeable; this keeps $\Psi_{\text{func}}$ compatible with future handoff paraphrase strategies.

\section{Additional Ablation Studies}
\label{app:additional-ablations}

This appendix complements the main-paper ablations of \S\ref{sec:ablation} with three pipeline-side studies: a synthesis-cost estimate (\S\ref{app:cost-study}), a cross-backbone fallback-rate analysis (\S\ref{app:fallback-study}), and a pipeline-component ablation on data-generation success rate (\S\ref{app:pipeline-ablation}).

\subsection{Synthesis Cost}
\label{app:cost-study}
We reran the same synthesis pipeline while recording API usage for 500 validated multi-turn trajectories. The measured average cost was \$0.49 per trajectory. Extrapolating this cost under the same pipeline configuration and trajectory mix gives an estimated cost of approximately \$4,410 for the released set of about 9k trajectories. This estimate covers synthesis-time API usage and excludes SFT training and downstream evaluation. Because comparable trajectory-level billing logs are unavailable for existing methods, we do not make a direct numerical cost comparison; local retry is expected to avoid some recomputation by refilling only failed calls rather than regenerating complete trajectories.

\subsection{Plan-Agent Fallback Rate Across Backbones}
\label{app:fallback-study}

To assess how faithfully different backbones honor the FSM-declared sources, we run $\mathcal{A}_{\text{plan}}$ on the same 500 FSM skeletons with four backbones and measure the \Pfallback{} rate, defined as the fraction of filled arguments for which the declared $s_0\in\{\Pinit,\Poutput,\Pumsg\}$ cannot be used directly and the planner records a recovery binding. A recovery uses a \Pinit{} or \Pcreate{} value when permitted; the resulting call is executor-validated after binding, and the original $s_0$ remains in \texttt{fallback\_from}. This is an \emph{argument-level} fallback (the \Pfallback{} Provenance Tag of \S\ref{sec:provenance}). We also report the distribution of declared sources actually consumed (\Pinit{} / \Poutput{} / \Pcreate{}). \Pumsg{} arguments are counted under \Pcreate{} in this distribution, because the remap-bind-restore protocol of Appendix~\ref{app:method-plan-pum} binds their values through the \Pcreate{} path at fill time and only restores the tag afterwards; the three source columns and Fb.\% therefore sum to 100 up to rounding. Lower fallback indicates stricter adherence to the Provenance Invariant. This study is a separate synthesis run from the pipeline ablation of Appendix~\ref{app:pipeline-ablation}, using its own sampled skeletons and initial configs, so the absolute proportions are not directly comparable across the two tables.

\begin{table}[t]
\centering\small
\setlength{\tabcolsep}{2.5pt}
\renewcommand{\arraystretch}{1.1}
\begin{tabular}{lrrrrrr}
\toprule
Backbone & Traj & Args & Fb.\% & \Pinit\% & \Poutput\% & \Pcreate\% \\
\midrule
Gemini-3.1 Pro            & 500 & 3{,}962 & \textbf{2.02} & 42.7 & 25.5 & 29.8 \\
Gemini-3 Pro            & 500 & 4{,}348 & 2.76 & 36.4 & 24.0 & 36.9 \\
DeepSeek-V4-Flash  & 500 & 3{,}253 & 3.08 & 28.3 & \phantom{0}9.2 & 59.4 \\
Qwen3.5-Plus       & 500 & 4{,}365 & 4.82 & 27.3 & 34.4 & 33.5 \\
\bottomrule
\end{tabular}
\caption{Cross-backbone \Pfallback{} study on identical FSM skeletons. Traj is the number of trajectories and Args the total number of filled arguments across them; Fb.\% is the runtime fallback rate; the remaining columns show the distribution of declared sources actually consumed by $\mathcal{A}_{\text{plan}}$.}
\label{tab:fallback-study}
\end{table}

\paragraph{Observations.} (i) Gemini-3.1 Pro achieves the lowest fallback (2.02\%) and the highest \Pinit{} utilization (42.7\%), indicating the strongest adherence to declared sources. (ii) Qwen3.5-Plus shows the highest fallback (4.82\%) with heterogeneous \texttt{fallback\_from} labels (\texttt{default}, \texttt{user\_request}, \texttt{user\_intent}), suggesting weaker compliance with the FSM-imposed source constraint. (iii) DeepSeek-V4-Flash exhibits an atypical distribution: \Poutput{} drops to 9.2\% while \Pcreate{} climbs to 59.4\%, and the total argument count is the lowest (3{,}253 vs.\ 4{,}348 for Gemini-3 Pro on the same skeletons). The backbone tends to fabricate literals rather than reference prior tool returns, producing shorter trajectories. (iv) Gemini-3 Pro sits in the middle (2.76\%) with all fallbacks routed through \Pcreate{}, giving the most predictable failure mode. Our main pipeline therefore stays within the Gemini family: Gemini-3.1 Pro drives $\mathcal{A}_{\text{FSM}}$, whose cost is amortized because one skeleton yields several sampled trajectories, while $\mathcal{A}_{\text{plan}}$ uses Gemini-3 Pro, whose endpoint remained stable at the request throughput a full synthesis run requires.

\subsection{Pipeline-Level Ablation}
\label{app:pipeline-ablation}

Table~\ref{tab:pipeline-ablation} reports the effect of turning off individual pipeline components on data generation success rate and argument-source distribution. All variants use Gemini-3 Pro as the backbone and share the same toolset and initial-config pool. This is an independent synthesis run from the cross-backbone study of Appendix~\ref{app:fallback-study} (different sampled FSM skeletons and initial configs), so the absolute source proportions differ from Table~\ref{tab:fallback-study}; only within-table comparisons across variants are meaningful. We generate 500 trajectories per variant and measure the executor pass-rate (Succ) and the fraction of arguments assigned to each source type (including the argument-level fallback rate \Pfallback), with \Pumsg{} again counted under \Pcreate{} as in Appendix~\ref{app:fallback-study}. In the Full, FSM+noTag, and noRetry variants, the FSM samples up to 3 paths per trajectory and each call can be refilled up to 3 times on executor failure. In the noFSM variants, only a single path is generated (no FSM path sampling), but the per-call 3-retry refill budget remains.

\begin{table}[t]
\centering\small
\setlength{\tabcolsep}{3pt}
\begin{tabular}{lccccc}
\toprule
% \textbf{Variant} & \textbf{Succ} & \textbf{Regen\%} & \textbf{\Poutput\%} & \textbf{\Pcreate\%} & \textbf{\Pinit\%} \\
\textbf{Variant} & \textbf{Succ} & \textbf{\Pfallback\%} & \textbf{\Poutput\%} & \textbf{\Pcreate\%} & \textbf{\Pinit\%} \\
\midrule
Full \cspg{}    & 89\% & 2.06 & 29.5 & 40.9 & 27.5 \\
FSM+noTag   & 92\% & 0.00 & 0.0  & 53.8 & 46.2 \\
noFSM+Tag   & 50\% & 15.75& 16.2 & 43.9 & 24.2 \\
noFSM+noTag & 66\% & 0.00 & 0.0  & 54.2 & 45.8 \\
noRetry     & 54\% & 3.45 & 24.8 & 42.3 & 29.4 \\
\bottomrule
\end{tabular}
\caption{Pipeline-level ablation. FSM = structure-first skeleton; Tag = Provenance Tag declarations; Succ = trajectories that pass executor validation; \Pfallback = argument-level fallback rate; remaining columns show the fraction of arguments assigned to each source type (\Pumsg{} counted under \Pcreate{}).}
\label{tab:pipeline-ablation}
\end{table}

Several patterns stand out. (i)~Removing the FSM skeleton (noFSM+Tag) nearly halves the success rate (89\% $\to$ 50\%) and drastically increases the argument-level fallback rate to 15.75\%. This confirms that the FSM's structure-first prior is the primary guarantor of causal chain existence: without the FSM's macro-level path planning, $\mathcal{A}_{\text{plan}}$ degenerates into free-form turn-by-turn generation. Although the historical context is preserved, the generated tool-call sequences misalign with the Tag-declared causal dependencies (i.e., a Tag declares a need for \Poutput{}, but the preceding tools fail to produce that specific return). Consequently, a massive number of declared \Poutput{} sources cannot be matched in the history, forcing the planner to fall back to \Pcreate{}. (ii)~Removing only the Provenance Tags (FSM+noTag) actually yields a slightly higher pass-rate (92\%) and zero fallback, but produces zero \Poutput{} arguments; the pipeline reverts to generating only local (\Pinit{} and \Pcreate{}) sources, losing all cross-turn dependency structure. (iii)~Disabling per-call retry (noRetry) cuts the success rate to 54\%, highlighting the importance of the executor-in-the-loop refill mechanism. (iv)~The noFSM+noTag variant confirms that the FSM and Tag components are complementary: without either, the pipeline produces only local arguments at a moderate pass-rate. Together, these results complement the training-level ablations of Table~\ref{tab:ablation} by showing that FSM, Tag, and Retry each contribute a distinct and necessary function at the data-generation stage.

\subsection{Out-of-Distribution Generalization on BFCL Single-Turn}
\label{app:ood-singleturn}

\cspg-4B is trained exclusively on multi-turn trajectories synthesized by the pipeline of \S\ref{sec:method}. To probe whether this multi-turn-focused training transfers to settings that differ in both interaction structure and task distribution, we evaluate \cspg-4B on the BFCL v4 single-turn track (Table~\ref{tab:ood-singleturn}), which is split into two complementary subsets:

\begin{itemize}[leftmargin=*,topsep=2pt,itemsep=1pt]
\item \textbf{Non-live} (simple, parallel, multiple, parallel\_multiple, irrelevance): static, BFCL-authored benchmark tasks. The distribution is controlled and reproducible; performance here primarily reflects basic function-calling capability (tool selection, argument filling, intra-turn parallelism).
\item \textbf{Live} (live\_simple, live\_parallel, live\_multiple, live\_parallel\_multiple, live\_relevance, live\_irrelevance): real-world user-contributed queries with noisier phrasing and broader intent coverage. Performance here is more sensitive to generalization and robustness.
\end{itemize}

\begin{table}[t]
\centering\small
\setlength{\tabcolsep}{6pt}
\renewcommand{\arraystretch}{1.05}
\begin{tabular}{lccc}
\toprule
\textbf{Split} & \textbf{Backbone} & \textbf{SAP-4B} & $\Delta$ \\
\midrule
Non-live & 84.36 & 84.88 & {\color{ForestGreen}+0.52} \\
Live & 75.41 & 75.95 & {\color{ForestGreen}+0.54} \\
\midrule
Overall avg & 79.89 & 80.42 & {\color{ForestGreen}+0.53} \\
\bottomrule
\end{tabular}
\caption{BFCL v4 single-turn results (accuracy, \%). Backbone is Qwen3-4B-Instruct-2507; \cspg-4B is the same backbone after SFT on \cspg{} data. Multi-turn-only training neither degrades single-turn performance nor regresses on the noisier Live split, and yields small consistent gains on both subsets.}
\label{tab:ood-singleturn}
\end{table}

Two takeaways. First, multi-turn-only training does not regress on single-turn tasks: Non-live accuracy edges up by $+0.52$, indicating that the multi-turn argument-provenance prior does not interfere with single-step tool selection or argument filling. Second, Live accuracy improves by a comparable $+0.54$; since Live queries are out-of-distribution with respect to both the synthesis-time tool schemas and the multi-turn structure, this constitutes a mild but consistent OOD generalization signal. The overall lift ($+0.53$) is modest in absolute magnitude but uniformly positive, suggesting that the structural prior induced by Provenance Tag constraints transfers as auxiliary signal even when the trajectory collapses to a single turn.

\section{Case Study: Local vs.\ Cross-Turn Argument Grounding}
\label{app:case-study}

To illustrate how \cspg{}'s provenance constraints shape trajectory structure, we present two simplified trajectories (Figures~\ref{fig:case-study-a} and~\ref{fig:case-study-b}). Trajectory~A uses only local sources; Trajectory~B contains a cross-turn \Poutput{} dependency. Table~\ref{tab:case-study} summarizes their dependency profiles.

\begin{table}[t]
\centering\small
\setlength{\tabcolsep}{3pt}
\begin{tabular}{lccccc}
\toprule
& \textbf{Turns} & \textbf{\Pinit} & \textbf{\Poutput} & \textbf{\Pcreate/\Pfallback} & \textbf{Span} \\
\midrule
Trajectory~A (local)      & 2 & 0 & 0 & 3 & 0 \\
Trajectory~B (cross-turn) & 3 & 0 & 2 & 1 & 2 \\
\bottomrule
\end{tabular}
\caption{Source distribution and max dependency span for the two case-study trajectories.}
\label{tab:case-study}
\end{table}

\begin{figure*}[t]
\centering
\begin{promptbox}{Trajectory~A: Local Grounding Only (Travel domain)}
\small
\noindent\textbf{Turn 1}\\[2pt]
\noindent\textit{User:} ``Show me all the credit cards saved on my travel account.''\\[2pt]
\noindent\textit{Assistant:}
\begin{Verbatim}
<tool_call>
[get_all_credit_cards()]
</tool_call>
<tool_response>
[{"credit_card_list": {"CARD001": {...},"CARD002": {...}}}]
</tool_response>
\end{Verbatim}
\vspace{4pt}

\noindent\textbf{Turn 2}\\[2pt]
\noindent\textit{User:} ``Now calculate how much 1.0 USD is in EUR.''\\[2pt]
\noindent\textit{Assistant:}
\begin{Verbatim}
<tool_call>
[compute_exchange_rate(base_currency="USD",target_currency="EUR", value=1.0)]
</tool_call>
<tool_response>
[{"exchanged_value": 0.91}]
</tool_response>
\end{Verbatim}
\vspace{2pt}
\noindent $\vartriangleright$ \texttt{base\_currency}, \texttt{target\_currency}, \texttt{value}: \Pcreate
\end{promptbox}
\caption{Trajectory~A with only local sources. The three argument slots of \texttt{compute\_exchange\_rate} are user-supplied literals (\Pcreate); the preceding zero-argument call contributes no argument slot, and no argument depends on a prior turn ($L{=}0$).}
\label{fig:case-study-a}
\end{figure*}

\begin{figure*}[t]
\centering
\begin{promptbox}{Trajectory~B: Cross-Turn \Poutput{} Dependency (Twitter domain)}
\small
\noindent\textbf{Turn 1}\\[2pt]
\noindent\textit{User:} ``First check whether I'm logged in, then show me a list of all the accounts I'm currently following.''\\[2pt]
\noindent\textit{Assistant:}
\begin{Verbatim}
<tool_call>
[posting_get_login_status(), list_all_following()]
</tool_call>
<tool_response>
[{"login_status": true}, {"following_list": ["dev_tips","popular_science", "movie_reviews", "ai_updates", "nature_news"]}]
</tool_response>
\end{Verbatim}
\vspace{3pt}

\noindent\textbf{Turn 2}\\[2pt]
\noindent\textit{User:} ``Now post a tweet with the text `Trying to tweet more intentionally...'.''\\[2pt]
\noindent\textit{Assistant:}
\begin{Verbatim}
<tool_call>
[post_tweet(content="Trying to tweet more intentionally...")]
</tool_call>
<tool_response>
[{"id": 16, ...}]
</tool_response>
\end{Verbatim}
\vspace{3pt}

\noindent\textbf{Turn 3}\\[2pt]
\noindent\textit{User:} ``Now mention all the accounts I follow in that tweet.''\\[2pt]
\noindent\textit{Assistant:}
\begin{Verbatim}
<tool_call>
[mention(tweet_id=16, mentioned_usernames=["dev_tips","popular_science", "movie_reviews", "ai_updates","nature_news"])]
</tool_call>
<tool_response>
[{"mention_status": "success"}]
</tool_response>
\end{Verbatim}
\vspace{2pt}
\noindent $\vartriangleright$ \texttt{content}: \Pcreate{} (Turn~2);\quad \texttt{tweet\_id}: \Poutput$(k'{=}2, j'{=}1, \texttt{id})$;\quad \texttt{mentioned\_usernames}: \Poutput$(k'{=}1, j'{=}2, \texttt{following\_list})$
\end{promptbox}
\caption{Trajectory~B with cross-turn \Poutput{} dependencies: Turn~3's \texttt{mention} consumes \texttt{tweet\_id} from Turn~2's \texttt{post\_tweet} return and \texttt{mentioned\_usernames} from Turn~1's \texttt{list\_all\_following} return (maximum span $L^\star{=}2$).}
\label{fig:case-study-b}
\end{figure*}

In Trajectory~A, every argument is a literal that the user supplies in the same turn that consumes it (\Pcreate); no cross-turn propagation occurs, and all arguments have $L=0$. In Trajectory~B, by contrast, the \texttt{mention} call consumes two cross-turn values: \texttt{tweet\_id} from Turn~2's \texttt{post\_tweet} return ($L=1$) and \texttt{mentioned\_usernames} from Turn~1's \texttt{list\_all\_following} return ($L=2$). \cspg{} produces these dependencies by design: the FSM skeleton declares the \Poutput{} edges at plan time, and $\mathcal{A}_{\text{plan}}$ resolves them against the real executor state before binding. Both trajectories are valid multi-turn dialogues, but \cspg{}'s provenance constraints make Trajectory~B the representative case, directly addressing the dependency-gap diagnostic of \S\ref{sec:prelim}.

% =====================================================================
\section{Prompt Templates}
\label{app:prompts}

This appendix lists condensed English templates of the four core prompts used by the three agents and the rewrite operators (Figures~\ref{fig:prompt-fsm}--\ref{fig:prompt-rewrite-hints}). Boilerplate (login etiquette, path conventions, output-format reminders) and few-shot examples are omitted; the released code carries the full prompts. Cyan placeholders such as \pv{N} are filled at synthesis time.

\begin{figure*}[t]
\centering
\begin{promptbox}{$\mathcal{A}_{\text{FSM}}$: FSM Skeleton Synthesis}
\textless task\textgreater\\
You are an FSM designer. Build executable, logically sound, diverse FSM skeletons for multi-turn tool-calling dialogs over a target toolset.\\
\textless /task\textgreater\\[3pt]
\textless structural\_constraints\textgreater\\
- Exactly one state with \texttt{type="initial"}; at least one terminal state.\\
- Each turn carries fields \texttt{from / to / action / function\_calls / prov\_tag}.\\
- \texttt{function\_calls} has length 1--10; names must come exactly from the provided tool list.\\
\textless /structural\_constraints\textgreater\\[3pt]
\textless design\_constraints\textgreater\\
- DAG-structured (no cycles); depth $\ge$ \pv{N} from $\sigma_0$.\\
- Call-count distribution: $\sim$40\% turns with 2--4 calls, $\sim$45\% with 5--7, $\sim$15\% with one. Each combination must reflect a realistic single-step user intent.\\
- Cross-turn dependencies: later turns may reference earlier returns or user-supplied values.\\
- Order: login first; search before result-dependent actions; status-check tools only in ``check'' intents.\\
- Long-tail coverage: at least one turn must use a tool from the emphasis subset \pv{emphasis\_set}.\\
\textless /design\_constraints\textgreater\\[3pt]
\textless provenance\_tag\_declaration\textgreater\\
For every required argument, declare \texttt{prov\_tag[tool][param]} from one of:\\
- \texttt{initial\_state}: value pre-exists in the initial-state summary (IDs, filenames, credentials).\\
- \texttt{prev\_output}: value read from a prior tool's return.\\
- \texttt{self\_create}: new user-supplied value (new filename, tweet text, search keyword).\\
- \texttt{prev\_user\_msg}: value already provided in an earlier user message.\\
Rules: every required param must appear; the first turn out of $\sigma_0$ may NOT use \texttt{prev\_output} or \texttt{prev\_user\_msg}; login credentials must use \texttt{initial\_state}.\\
\textless /provenance\_tag\_declaration\textgreater\\[3pt]
\textless inputs\textgreater\\
Available tools: \pv{tool\_summaries}\\
Initial-config summary: \pv{initial\_state}\\
\textless /inputs\textgreater\\[3pt]
\textless output\textgreater\\
Output ONLY \texttt{\textless fsm\textgreater JSON\textless /fsm\textgreater} matching \pv{schema\_hint}. No explanatory text.\\
\textless /output\textgreater
\end{promptbox}
\caption{Prompt template for $\mathcal{A}_{\text{FSM}}$ (FSM skeleton synthesis).}
\label{fig:prompt-fsm}
\end{figure*}

\begin{figure*}[t]
\centering
\begin{promptbox}{$\mathcal{A}_{\text{plan}}$: Per-Call Argument Filling}
\textless task\textgreater\\
You are filling parameters for a SINGLE tool call at turn \pv{turn\_idx}. Tool to call: \pv{func\_name}.\\
\textless /task\textgreater\\[3pt]
\textless inputs\textgreater\\
- Tool signature: description, required params, properties: \pv{tool\_doc}\\
- Declared Provenance Tags from Phase 1 FSM: \pv{prov\_tag}\\
- Previous turns history (executed calls and their returns): \pv{history}\\
- Initial state (starting state; may be modified by earlier turns): \pv{initial\_state}\\
- Last attempt error (if refill): \pv{last\_error}\\
\textless /inputs\textgreater\\[3pt]
\textless output\_format\textgreater\\
Output \texttt{\textless result\textgreater JSON\textless /result\textgreater} with two keys:\\
- \texttt{args\_for\_exec}: the actual parameter values for execution. \texttt{null} is only acceptable for OPTIONAL params whose documentation explicitly states that \texttt{null} means ``no filter / default behavior''.\\
- \texttt{args\_provenance}: per-argument provenance metadata. \texttt{src} MUST be one of:\\
\hspace*{1em}\textbullet\ \texttt{initial\_state}: value from initial state (also set \texttt{config\_path})\\
\hspace*{1em}\textbullet\ \texttt{prev\_output}: value from a prior turn's tool return (also set \texttt{ref\_turn} and \texttt{ref\_field})\\
\hspace*{1em}\textbullet\ \texttt{self\_create}: value you create in this turn\\
\hspace*{1em}\textbullet\ \texttt{prev\_user\_msg}: value the user provided in an earlier turn (also set \texttt{introduce\_in\_turn} $<$ current turn)\\
\hspace*{1em}\textbullet\ \texttt{fallback}: declared upstream is unavailable and this binding uses a recovery value; the resulting call must then be validated by execution (also set \texttt{fallback\_from} to the original src)\\
\textless /output\_format\textgreater\\[3pt]
\textless rules\textgreater\\
- If the declared source is unavailable, recover the argument from \Pinit{} or \Pcreate{} when permitted. Record \texttt{fallback} whenever this recovery path replaces the declared source. Bind the recovery value and let the executor validate the resulting call.\\
- NEVER fabricate existing entities or credentials: IDs, tokens, usernames, passwords, or existing resource paths must be grounded in an executor-visible source.\\
- For \texttt{prev\_user\_msg}, \texttt{introduce\_in\_turn} must be strictly before the current turn.\\
- For \texttt{prev\_output}, \texttt{ref\_turn}/\texttt{ref\_field} must identify a real prior return field.\\
\textless /rules\textgreater
\end{promptbox}
\caption{Prompt template for $\mathcal{A}_{\text{plan}}$ (per-call argument filling). The two-track output $(\boldsymbol{\theta}^{\text{exec}}, \boldsymbol{\theta}^{\text{prov}})$ corresponds to \texttt{args\_for\_exec} and \texttt{args\_provenance}.}
\label{fig:prompt-plan}
\end{figure*}

\begin{figure*}[t]
\centering
\begin{promptbox}{$\mathcal{A}_{\text{msg}}$: User-Message Synthesis}
\textless task\textgreater\\
You are a multi-turn dialogue data generator. Given a tool-calling sequence (GT) for the current turn, generate the corresponding natural user message.\\
\textless /task\textgreater\\[3pt]
\textless inputs\textgreater\\
- Prior conversation history (the user said what, the assistant called which tools, what they returned): \pv{prior\_text}\\
- Current turn's expected tool calls (GT): \pv{turn\_gt}\\
- Per-argument source guide for this turn (the must-mention list \pv{must\_mention}, marker semantics below).\\
\textless /inputs\textgreater\\[3pt]
\textless must\_mention\_markers\textgreater\\
Express each parameter according to its source marker:\\
- \texttt{[NEW]}: explicitly state the value (e.g., filename, number, keyword).\\
- \texttt{[FROM\_HISTORY]}: use a vague reference (``that file'', ``the ID you found earlier''); do NOT repeat the exact value, but the reference must be unambiguous.\\
- \texttt{[FROM\_CONFIG]}: explicitly state the exact value verbatim. The assistant cannot read the initial state, so IDs, credentials, and token-like fields must appear character-for-character.\\
- \texttt{[FALLBACK\_from\_$\ldots$]}: the original source was unavailable and a value was created; treat as \texttt{[NEW]} and state it explicitly.\\
\textless /must\_mention\_markers\textgreater\\[3pt]
\textless core\_constraints\textgreater\\
1. \textbf{Assistant capability boundary}: the assistant can ONLY call tools, not reply with text. The user message is the only input driving its behavior; it must contain enough information for correct tool calls.\\
2. \textbf{Parameter completeness}: every parameter used this turn must be naturally expressed in the user message according to its marker.\\
3. \textbf{Semantic alignment}: the user message's semantic direction must match what the tools actually do (e.g., \texttt{unfollow} $\ne$ ``follow'', \texttt{rm} = ``delete'').\\
4. \textbf{Intent-shift handling}: if this turn's intent differs from prior turns, explicitly state the transition.\\
5. \textbf{History is fact}: write the message conditioned on the provided history; do not modify history.\\
6. \textbf{Full GT coverage}: if the GT lists $n$ tool calls, the user message must naturally cover \emph{every} call (chain them with ``first \ldots, then \ldots, finally \ldots'' when needed).\\
\textless /core\_constraints\textgreater\\[3pt]
\textless style\textgreater\\
60\% conversational + 40\% instructional; 1--3 sentences; named values quoted; never reference tool internals.\\
\textless /style\textgreater\\[3pt]
\textless output\textgreater\\
\texttt{\textless msg\textgreater\{"turn\_idx": \pv{turn\_idx}, "user\_msg": "\ldots"\}\textless /msg\textgreater}\\
\textless /output\textgreater
\end{promptbox}
\caption{Prompt template for $\mathcal{A}_{\text{msg}}$ on user-message synthesis. The must-mention list is pre-rendered by the dispatcher of Appendix~\ref{app:method-msg} from $\boldsymbol{\theta}^{\text{prov}}$ and $\Delta_v$, with each entry already tagged \texttt{[NEW]} / \texttt{[FROM\_HISTORY]} / \texttt{[FROM\_CONFIG]} / \texttt{[FALLBACK from $\ldots$]}.}
\label{fig:prompt-msg-user}
\end{figure*}

\begin{figure*}[t]
\centering
\begin{promptbox}{$\mathcal{A}_{\text{msg}}$: Assistant Text Response (on no-tool-call turn)}
\textless task\textgreater\\
You are the assistant in a multi-turn tool-calling conversation. Write the NEXT assistant reply body when this turn has no further tool calls.\\
\textless /task\textgreater\\[3pt]
\textless requirements\textgreater\\
1. Output ONLY natural language, no tool calls, no JSON, no placeholders.\\
2. If tasks have been completed, briefly summarize and ask if anything else is needed.\\
3. If the task cannot continue (missing tool, missing argument, permission), explain why politely and suggest next steps.\\
4. Match the language of the user.\\
5. Report errors honestly: never claim success when prior tool returns contain \texttt{error / fail / exception / empty / NULL}.\\
6. NEVER answer the requested tool-action from general knowledge, NEVER invent a result, NEVER claim the action was done unless prior tool messages prove it.\\
7. Blocking-reply rule: if later turns may supply more tools or missing arguments, this current reply must state what is missing rather than pre-computing the answer.\\
\textless /requirements\textgreater
\end{promptbox}
\caption{Prompt template for $\mathcal{A}_{\text{msg}}$ on assistant text responses (turns with $\mathcal{C}_k = \varnothing$).}
\label{fig:prompt-msg-assistant}
\end{figure*}

\begin{figure*}[t]
\centering
\begin{promptbox}{Rewrite-Operator Hints ($\Psi_{\text{param}}$, $\Psi_{\text{func}}$)}
\textless MISS\_PARAM\_CLOSED\_LOOP\textgreater\\
\textit{(injected into $\mathcal{A}_{\text{plan}}$ at the recovery turn $k^*+1$)}\\
The previous turn $k^*$ should have called \pv{tool} but lacked required parameters \pv{missing\_params}. The current turn's user message has supplied them. Call \pv{tool} now with all required parameters (including the previously missing ones), unless the user's intent has clearly shifted, in which case call a more appropriate tool. Logical alignment with the user takes priority over closing the loop.\\
\textless /MISS\_PARAM\_CLOSED\_LOOP\textgreater\\[6pt]
\textless TOOL\_INJECTION\textgreater\\
\textit{(injected at the recovery turn $k^*+1$)}\\
The current turn is a tool-injection turn: the system has just registered new tools \pv{func\_names} into the visible toolkit. Prefer these newly registered tools when they match the user's intent. Even when \pv{question\_k} is the empty list \texttt{[]}, the ground truth must remain non-empty: extract the required parameters from the prior turn $k^* = k - 1$, whose unmet request triggered the injection.\\
\textless /TOOL\_INJECTION\textgreater\\[6pt]
\textless MISSING\_FUNCTION\_REQUEST\textgreater\\
\textit{(injected at the request turn $k^*$, before the handoff message)}\\
The current turn is a request turn: the user asks for tools \pv{func\_names} that are NOT yet in the visible function-doc. You MUST output \texttt{[]} for this turn, even if other tools are available; calling any tool here breaks the operator's semantics.\\
\textless /MISSING\_FUNCTION\_REQUEST\textgreater
\end{promptbox}
\caption{Hint templates injected by the rewrite operators $\Psi_{\text{param}}$ and $\Psi_{\text{func}}$. Placeholders are filled at injection time.}
\label{fig:prompt-rewrite-hints}
\end{figure*}

\end{document}